\documentclass{article}

\usepackage[final]{neurips_2021}

\usepackage[utf8]{inputenc} 
\usepackage[T1]{fontenc}    
\usepackage{hyperref}       
\usepackage{url}            
\usepackage{booktabs}       
\usepackage{amsfonts}       
\usepackage{amsmath}       
\usepackage{nicefrac}       
\usepackage{microtype}      
\usepackage{xcolor}         
\usepackage{latexsym}

\usepackage{graphicx}
\usepackage{booktabs}
\usepackage{multirow}
\usepackage{float}
\usepackage{diagbox}
\usepackage{arydshln}
\setcitestyle{round}
\usepackage[export]{adjustbox}
\usepackage{refcount}
\usepackage{wrapfig}
\usepackage{amssymb}

\usepackage{microtype}
\usepackage{subcaption}
\usepackage{xargs}                    
\usepackage[colorinlistoftodos,prependcaption,textsize=tiny]{todonotes}
\newcommandx{\angeliki}[2][1=]{\todo[linecolor=red,backgroundcolor=red!25,bordercolor=red,#1]{#2}}
\newcommandx{\adhi}[2][1=]{\todo[linecolor=red,backgroundcolor=red!25,bordercolor=blue,#1]{#2}}
\newcommandx{\egribovskaya}[2][1=]{\todo[linecolor=blue,backgroundcolor=blue!25,bordercolor=blue,#1]{#2}}
\newcommandx{\tayfunterzi}[2][1=]{\todo[linecolor=yellow,backgroundcolor=yellow!25,bordercolor=yellow,#1]{#2}}
\newcommandx{\maigimenez}[2][1=]{\todo[linecolor=pink,backgroundcolor=pink!25,bordercolor=pink,#1]{#2}}
\newcommandx{\seb}[2][1=]{\todo[linecolor=olive,backgroundcolor=olive!25,bordercolor=olive,#1]{#2}}
\newcommandx{\phil}[2][1=]{\todo[linecolor=green,backgroundcolor=green!25,bordercolor=green,#1]{#2}}
\newcommandx{\adam}[2][1=]{\todo[linecolor=purple,backgroundcolor=purple!25,bordercolor=purple,#1]{#2}}

\newcommandx{\arxiv}{\textbf{\textsc{arXiv}}}
\newcommandx{\wmt}{\textbf{\textsc{WMT}}}
\newcommandx{\customnews}{\textbf{\textsc{CustomNews}}}
\newcommandx{\timestratified}{{\textsc{time-stratified}}}
\newcommandx{\control}{{\textsc{control}}}
\newcommandx{\LFTRHFTE}{{\textsc{emerging new words}}}
\newcommandx{\datalink}{{\url{https://github.com/deepmind/deepmind-research/tree/master/pitfalls_static_language_models}}}
\newcommandx{\anonymouslink}{{\url{https://anon.ymous}}}
\newcommand{\ignore}[1]{}
\newcommand{\dyignore}[1]{}

\newenvironment{itemizesquish}{\begin{list}{\labelitemi}{\setlength{\itemsep}{0em}\setlength{\labelwidth}{0.5em}\setlength{\leftmargin}{\labelwidth}\addtolength{\leftmargin}{\labelsep}}}{\end{list}}

\title{Mind the Gap: Assessing Temporal Generalization in Neural Language Models}

\ignore{\author{%
  David S.~Hippocampus\thanks{Use footnote for providing further information
    about author (webpage, alternative address)---\emph{not} for acknowledging
    funding agencies.} \\
  Department of Computer Science\\
  Cranberry-Lemon University\\
  Pittsburgh, PA 15213 \\
  \texttt{hippo@cs.cranberry-lemon.edu} \\
}}

\author{
   Angeliki Lazaridou\thanks{Equal contribution. $^\spadesuit$ Project initiation.
   $^\triangle$ Paper writing.
   $^\diamondsuit$ Project technical infrastructure.
   $^\heartsuit$ Model design and experiments.
   $^\clubsuit$ Project support and advice.}~~$^{\heartsuit\triangle\spadesuit}$ Adhiguna Kuncoro$^{\star\heartsuit\triangle}$ Elena Gribovskaya$^{\star\heartsuit\triangle}$ \\
   \textbf{Devang Agrawal}$^{\diamondsuit\heartsuit}$ \textbf{Adam Li\v{s}ka}$^{\diamondsuit\heartsuit}$ \textbf{Tayfun Terzi}$^{\diamondsuit}$ \textbf{Mai Gimenez}$^{\diamondsuit}$
   \\ \textbf{Cyprien de Masson d'Autume}$^{\diamondsuit}$ \textbf{Tomas Kocisky}$^{\heartsuit}$ \textbf{Sebastian Ruder}$^{\heartsuit}$\\\textbf{Dani Yogatama}$^{\clubsuit}$ \textbf{Kris Cao}$^{\clubsuit}$  \textbf{Susannah Young}$^{\clubsuit}$ \textbf{Phil Blunsom}$^{\clubsuit\spadesuit}$
   \\
   DeepMind, London, UK \\
   {\small \tt \{angeliki,akuncoro,egribovskaya\}@deepmind.com} \\
}

\begin{document}

\maketitle

\setcounter{footnote}{0}

\begin{abstract}
Our world is open-ended, non-stationary, and constantly evolving; thus what we talk about and how we talk about it change over time. This inherent dynamic nature of language contrasts with the current static language modelling paradigm, which trains and evaluates models on utterances from overlapping time periods. Despite impressive recent progress, we demonstrate that Transformer-XL language models perform worse in the realistic setup of predicting future utterances from beyond their training period, and that model performance becomes increasingly worse with time. We find that, while increasing model size alone---a key driver behind recent progress---does not solve this problem, having models that continually update their knowledge with new information can indeed mitigate this performance degradation over time. Hence, given the compilation of ever-larger language modelling datasets, combined with the growing list of language-model-based NLP applications that require up-to-date factual knowledge about the world, we argue that now is the right time to rethink the static way in which we currently train and evaluate our language models, and develop \emph{adaptive} language models that can remain up-to-date with respect to our ever-changing and non-stationary world. We will publicly release our dynamic, streaming language modelling benchmarks for \wmt~and \arxiv~to facilitate language model evaluation that takes temporal dynamics into account.\footnote{We release our dynamic (streaming) language modelling benchmark for \wmt~and \arxiv~at~\datalink.} 
\end{abstract}

\section{Introduction}
\label{sec:intro}
In recent years, substantial efforts in neural language modelling have focused on finding better neural architectures, building increasingly larger models, and compiling ever-larger amounts of training data, which have been shown to endow language models with the ability to perform well on a wide variety of downstream tasks with minimal fine-tuning~\citep{Vaswani:etal:2017,Radford:etal:2019,Brown:etal:2020}. While this approach has led to impressive progress, it nevertheless relies on a static experimental paradigm. Concretely, the prevailing practice is to curate a large pretraining web crawl---randomly partitioned into a training set and a validation set in a time-agnostic fashion---and then evaluate on tasks and benchmarks that mostly overlap in time with the pretraining data.\footnote{In the case of GPT-3 \citep{Brown:etal:2020}, such tasks include LAMBADA \citep{lambada}, TriviaQA \citep{triviaqa}, and WMT translation datasets, among others. These tasks were introduced between 2014 and 2017, which overlap in time with the GPT-3 CommonCrawl dataset that covered the period of 2016-2019.}

 In this work, we argue that such practices carry two potential risks. First, they do not assess a language model's ability to generalize well to future data from beyond their training period---an important ability we henceforth refer to as temporal generalization. In our dynamic and non-stationary world, temporal generalization is a key necessity: Many practical machine learning systems that use language model (LM) pretraining, such as machine translation and dialogue systems, are deployed on utterances that users will say in the future, whilst being trained on utterances that users have already said in the past. Furthermore, temporal generalization is also crucial to perform well on realistic use cases of language models in the real world. Examples include flagging fake news about recent events that happen outside of the training period \citep{thorne-vlachos-2018-automated,zellers_etal_2019,Augenstein2019}, forecasting stock prices from the latest news articles \citep{ding_2015}, and answering knowledge-intensive questions like ``How many people have been infected by COVID-19?'' and ``Has the USA ever had a female Vice President?'', whose answers have evolved with time. 
 
 Second, the temporal overlap between the training and evaluation data increases the risk of ``test data contamination'', where parts of the evaluation task are unwittingly included in the pretraining data. Indeed, many language modelling evaluations treat the data as independent and identically distributed (i.i.d) at either the sentence \citep{Chelba:etal:2013} or document level \citep{Brown:etal:2020, Gao:etal:2021}. Nevertheless, language modelling data are not i.i.d. (neither at the word, sentence, or document level); rather it is a time series, and thus models trained on the prefix of a sample from the series should be evaluated on the continuation of that series. While previous research~\citep{levenberg_2010} has highlighted the importance of temporal splits for fairer and more realistic evaluations---and has led to research~\citep{Osborne:VanDurme:2014, Yogatama:etal:2014} that addresses language modelling from this streaming perspective (\S\ref{sec:background})---using temporal splits (or splits beyond random ones) is still the exception rather than the rule, as evidenced by many contemporary LM~\citep{Brown:etal:2020, Gao:etal:2021} and downstream tasks~\citep{Lewis:etal:2020b} that are affected by test data contamination.\footnote{\cite{Brown:etal:2020} used $n$-gram filtering and deduplication to remove overlaps between the training and test sets. This can potentially induce a correlation between the training and evaluation sets that LMs can exploit.}  

Here we begin with our first question: To what extent does the current static language modelling practice overestimate performance, compared to the more realistic setup that evaluates LMs on future utterances? To this end, we introduce our dynamic, streaming language modelling benchmarks (\S\ref{sec:experimental}), and find that Transformer-XLs \citep{Dai:etal:2019} perform up to 16\% worse when predicting articles that are published up to 2 years after the end of the training period. Moreover, model performance becomes increasingly worse with time (\S\ref{sec:overall_ppl}). Given this finding, we ask: What kinds of predictions is the model struggling with in the dynamic evaluation setup?----which we answer in \S\ref{sec:analysis}. 

Beyond LM perplexity evaluation, we further ask: How exactly does this temporal performance degradation of Transformer LMs manifest in different types of question-answering (QA) tasks? We answer this through two different QA tasks, including one around recent events happening outside of the LM training period (\S\ref{sec:downstream}). Lastly, given the challenges presented by temporal generalization for LMs: What, then, is the remedy? This question is important because keeping LMs up-to-date by retraining with new data is expensive in compute and carbon costs \citep{strubell-2019,carbon-impact-2021}, and risks the model getting outdated in-between long retraining cycles.\footnote{These risks are exacerbated by the trend of ever-larger LMs, where retraining incurs even higher costs. 
} We find that increasing model size alone---a key driver behind recent LM progress \citep{Kaplan:etal:2020}---is not a solution for the temporal generalization problem (\S\ref{sec:model_size}): Larger models suffer from the same performance degradation with time, and a smaller model trained on more recent data can outperform a 60\% larger model that lacks access to more recent data.  
We then explore a simple yet effective way of keeping our models up-to-date by continually updating the model's parameters through dynamic evaluation \citep{mikolov_etal_2010, Krause:etal:2019}, which performs a few steps of gradient descent on streams of new data (\S\ref{sec:solution}), and outline other promising approaches in this direction (\S\ref{sec:background}).
We conclude with the following recommendations for future LM research:
\begin{itemizesquish}
    \item We should evaluate LMs on their generalization ability to future data, which circumvents test data contamination, rewards models that generalize beyond the surface patterns of their pretraining data, and better reflects how large LMs are used in practical systems. We thus argue for the broader inclusion of timestamp information in pretraining data and downstream tasks to make this possible.
    \item Stale LMs that are deployed far outside of their training period perform substantially worse on downstream tasks that require up-to-date factual knowledge, although a broader set of experiments are needed to pinpoint what kinds of tasks are most affected. Our findings also highlight the need for more tasks, benchmarks, and metrics that evaluate \emph{how well} and \emph{how rapidly} LMs are able to integrate new information, which are important ingredients to encourage progress in this direction.
    \item All in all, above and beyond impressive scaling efforts towards ever-larger models~\citep{Brown:etal:2020,Fedus:etal:2021}, we argue for the development of \emph{adaptive} language models that can remain up-to-date with respect to our open-ended and non-stationary world.
\end{itemizesquish}

\vspace{-4mm}
\section{Time-stratified language modelling}
\label{sec:experimental}

We begin by introducing our time-stratification experimental setup, which examines \emph{how well} Transformer LMs perform when evaluted on future utterances from beyond their training period.

\subsection{Datasets}
\label{sec:data}

We identify news and scientific articles as two sources of dynamic streaming data with a naturally changing distribution over time---lending themselves well to evaluating how well language models generalize over time. For the scientific domain, we use the publicly available arXiv abstracts (\arxiv).\footnote{\label{foot:links}ArXiv: \url{https://arxiv.org/help/oa/index}; WMT News: \url{http://data.statmt.org/news-crawl}; and SacreMoses: \url{https://github.com/alvations/sacremoses}. 
} For news, we use the publicly available WMT News Crawl (\wmt).\footnotemark[\getrefnumber{foot:links}] We ensure that any trends we observe also generalize well to models trained on larger datasets---which reliably improve language modelling and downstream task performance \citep{roberta}---by compiling a larger news corpus that we term \customnews. This dataset consists of crawled English news sources from 1969-2019, and covers various topics including politics, finance, and sport. 
We apply minimal preprocessing through: (i) Removal of non-English documents, (ii) deduplication using 
the MinHash algorithm, and (iii) tokenization using Moses.\footnotemark[\getrefnumber{foot:links}] Table~\ref{tab:dataset_statistics} summarizes key statistics of our datasets.

\begin{table*}
    \centering
     \resizebox{0.96\textwidth}{!}{%
    \begin{tabular}{c|c|r|r||r|r}
        \textbf{Dataset}&\textbf{Domain}& \textbf{Time period}
             & \textbf{\shortstack{\#Words per Doc\\ (Average)}} & \textbf{\shortstack{Training Size\\ (in GB)}}
            &\textbf{\shortstack{Prop. of \control's  \\Training Data\\ from the Test Period}}\\ \hline
        \wmt~&News  &2007 - 2019   &551 &22.65 &6.3\%  \\ 
        \customnews~&News &1969 - 2019    & 491 &395.59 &34.8\%\\ 
        \arxiv~ &Scientific text  &1986 - 2019  & 172 & 0.72 &14.5\% \\ 
    \end{tabular}}
    \caption{Statistics and time periods of the datasets used in this study.}
    \label{tab:dataset_statistics}
\end{table*}

\subsection{Experiment: A model up to 2 years stale}
\label{sec:setup}

\paragraph{Evaluation period and test set.} For each dataset, we pick the last two years (i.e. 2018 and 2019) as our evaluation period, and sub-sample a test set of 24k test documents (1k per test month).
\vspace{-4mm}
\paragraph{\timestratified~setup.} 
In this setup, we evaluate LMs trained on the past based on their ability to predict future articles that are published after the time period of their training data; this split is constructed using the time stamp of each article. Here we use all documents from the beginning of each dataset's time period up until September 2017 as training data, and use the last three months of 2017 as our validation period; 
we denote this as the \textbf{\timestratified}~setup. We then evaluate the model on the 2018-2019 test set above, which evaluates the model's ability to generalize across time by predicting articles up to two years after the end of their training period---a realistic time frame during which we expect large-scale language models to be used without retraining on recent data.

\vspace{-4mm}
\paragraph{\control~setup.} We assess whether time stratification 
poses a challenge for current LMs by comparing it with the following \textbf{\control}~setup. In this setup, the training set includes documents that come from the same 2018-2019 period as the evaluation set (naturally excluding the test documents themselves). This \control~setup thus resembles the prevailing (static) language modelling experimental practices, which train and evaluate LMs on text data from overlapping time periods.

Crucially, we control such that the two training sets are of the exact same size, i.e., they differ only in the time periods of their training data, rather than in their absolute training set sizes. Here we construct the \control~training data by taking the most recent documents starting from the end of the evaluation period (excluding the test documents and including the same number of training documents per test month), and keep adding documents from previous time periods until we reach the same training size as the \timestratified~setup. In Table~\ref{tab:dataset_statistics}, we report the proportion of documents in the \control~setup's training data that come from the same 2018-2019 time period as the evaluation set,
which is higher for \arxiv~and \customnews~due to their recent exponential growth of new documents. We sample a similarly-sized validation set as the \timestratified~setup, which in this case comes from the 2018-2019 evaluation period (again excluding the test documents). Importantly, both the \timestratified~and \control~models are evaluated on the exact same test set from the 2018-2019 period, which facilitates a fair perplexity comparison between the two setups.

\vspace{-4mm}
\paragraph{Relative perplexity comparison.} We want to measure temporal degradation, i.e. do Transformer LMs perform increasingly worse when predicting test documents further into the future? However, any \emph{absolute} perplexity degradation of the \timestratified~model over time (e.g., perplexity for Jan. 2018 vs Dec. 2018) is an unreliable measure: Some months have longer documents, which lead to higher perplexity. We thus measure temporal degradation through \emph{relative} perplexity changes between the \timestratified~and \control~models for the same test month (e.g. Dec. 2018).

\subsection{Model}
\label{sec:model}

We perform our experiments on autoregressive, left-to-right LMs. We use a Transformer-XL~\citep{Dai:etal:2019} with 18 layers and 1,024 hidden units, resulting in 287M parameters---roughly 15\% smaller than GPT-2$_{\text{MEDIUM}}$ and BERT$_{\text{LARGE}}$; we later explore larger models in \S\ref{sec:model_size}. We set the Transformer sequence length to 1,024, and set the memory cache length to 384 during training and 1,600 during test. We use a vocabulary of 50,259 subwords, obtained via SentencePiece~\citep{Kudo:Richardson:2018} trained on a random subset (up to 15GB) of the training data of each respective experiment, i.e., \control~and \timestratified.
Training and validation are done on subword tokens, but to facilitate our later analysis (\S\ref{sec:analysis}), all test perplexities are computed over actual test word tokens,\footnote{An example is detokenizing ``\emph{\_\_contact}'', ``\emph{less}'' into ``\emph{contactless}'', where ``\_\_'' denotes a token boundary.} whose negative log probabilities are obtained by summing those of their subwords.
\section{Language Modelling Experiments \& Analysis}
\label{sec:experiment1}
\begin{wraptable}{r}{0.5\textwidth}
 \centering
 \vspace{-15pt}
\resizebox{0.5\textwidth}{!}{%
\begin{tabular}{l  c c c}
\textbf{Setup} & \wmt & \shortstack{\textbf{\textsc{Custom}}\\\textbf{\textsc{News}}}&  \arxiv \\
    \control & 21.11 & 18.38 & 21.38\\
    \timestratified & 22.45 & 21.33 & 23.07\\\hline\hline
    $\Delta$, absolute & +1.34 & +2.95  & +1.69 \\
    $\Delta$, relative (\%) & 6.34 & 16.04 & 7.90 \\
\end{tabular}}
\caption{Perplexity of Transformer-XL when trained with the two different setups, and evaluated on the same test set from the 2018-2019
period.\label{tab:overall_perplexity}}
 \vspace{-5pt}
\end{wraptable}

\textbf{To what extent does the static \control~setup overestimate model performance, compared to the more realistic \timestratified~setup that evaluates LMs on future utterances?}
\label{sec:overall_ppl}
\begin{wrapfigure}{r}{0.5\textwidth}
\vspace{-20pt}
\includegraphics[width=8cm,height=5cm,keepaspectratio]{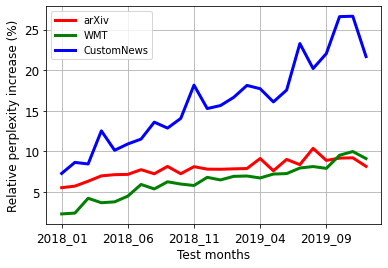}
\caption{Relative ppl. increase of \timestratified~over \control, across test months.\label{fig:over_month_all}}
 \vspace{-20pt}
\end{wrapfigure}
Figure \ref{tab:overall_perplexity} presents the results of our first experiment. Although we train both models: (i) On the exact same dataset sizes, and (ii) using the same model architectures, a stale \timestratified~model performs worse than the \control~model, which \emph{has} seen training data from the test period---with up to 16\% perplexity difference. We attribute the higher relative degradation on \customnews~and \arxiv~to their recent exponential growth of new documents, resulting in a higher proportion of documents from the test period in the data (Table~\ref{tab:dataset_statistics}), hence presenting a more difficult temporal generalization problem.

\textbf{Do Transformer LMs perform increasingly worse when predicting future utterances further away from their training period?} To this end, Fig.~\ref{fig:over_month_all} plots the relative perplexity increase of the \timestratified~over the \control~model. As evidenced by the upward slope on all datasets, the model deteriorates more as we ask it to predict data further away from the training period, affirming that the model indeed becomes \emph{increasingly outdated} with time. 
\textbf{How general are these findings?} We find that the same patterns not only generalize across datasets, as we have just shown, but are also found: (i) For test years other than 2018-2019 (Appendix \ref{sec:rolling}), (ii) beyond the two-year temporal gap between the end of the training and test periods (Appendix \ref{sec:increasing}), and (iii) across other languages (German WMT, Appendix \ref{sec:german}).

\subsection{Analysis}
\label{sec:analysis}
\begin{wrapfigure}{R}{0.5\textwidth}
    \centering
    \includegraphics[width=9cm,height=5cm,keepaspectratio]{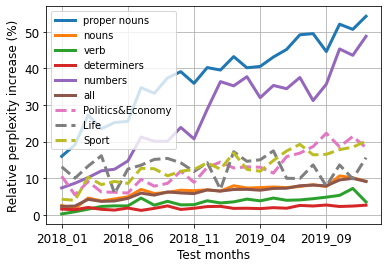}
\caption{\wmt~relative ppl. increase of the \timestratified~over the \control~models, broken down by part-of-speech (POS) tags (solid lines) and topics (dotted lines).}
\label{fig:overtime_pos}
\end{wrapfigure}

Having established that model performance degrades with time, we now turn to investigate the following question: What exactly are the kinds of predictions that the model is struggling with?
\vspace{-4mm}
\paragraph{Part-of-speech (POS) tag breakdown.} We present the relative perplexity increase of  the \timestratified~over the \control~model, broken down by POS tag and across time (Fig.~\ref{fig:overtime_pos}, solid lines). First, we see that performance on common nouns (orange line), the most frequent POS tag, degrades with time; in fact, performance degradation on common nouns drives the overall degradation trend (brown line). Moreover, the \timestratified~model's performance degrades most rapidly when making temporal generalizations about proper nouns (blue line) and numbers (purple line). 
Qualitative analysis indicates that the model performs badly on named entities in politics, whose position changed during our 2018-2019 evaluation period (e.g.,  ``Bolsonaro'', ``Pompeo'',  ``Khashoggi''). This degradation is consequential because proper nouns---and by extension named entities---closely relate to up-to-date factual world knowledge; in \S\ref{sec:downstream} we explore how exactly this degradation affects different downstream tasks. Interestingly, we also found the model struggling with concepts associated with cultural and sociological changes on which public perception and discourse have evolved over time,  such as ``MeToo'' and ``BlackLivesMatter''~\citep{Bender:etal:2021}. 

\vspace{-4mm}

\paragraph{Perplexity and topics.} We analyze \emph{how} the speed of the \timestratified~model's perplexity degradation relates to different topics. We first cluster the documents using Latent Dirichlet Allocation \cite[LDA]{Blei_lda}, which represents each document as a mixture of topics and each topic as a distribution over words; we then aggregate the perplexity of words in the test documents by topic. We observe that model performance on topics around politics and sports change more rapidly with time than topics around lifestyle, as shown in (Fig.~\ref{fig:overtime_pos}, shown in the three dotted lines).
\label{sec:topics}

\vspace{-4mm}
\paragraph{Perplexity and temporal frequency shifts } 
\label{sec:adaptation}
In practice, \emph{adaptation} is a key necessity to maximize the potential of LMs in our dynamic and non-stationary world. This includes the ability to integrate information about new words and concepts that \emph{never occurred} in the past, and also words whose context or meaning \emph{had substantially changed} across time. This need is well-reflected in our datasets: About 27\% of word types (i.e. unique words) on \customnews~each month had never occurred in the training period, such as ``Brexiteers'' and ``MeToo''.  We refer to these as \LFTRHFTE, and argue that these concepts are important because they reflect precisely the dynamic nature of our non-stationary world. Perhaps the most notable recent \LFTRHFTE~is ``COVID-19'', which had zero unigram probability prior to late-2019, and yet constitutes an important use case of the NLP systems today.
\begin{wraptable}{R}{0.5\textwidth}
\vspace{-10pt}
\resizebox{0.48\textwidth}{!}{
\begin{tabular}{c c |c c}  \toprule
\multicolumn{2}{c|}{\diagbox{\shortstack{Occur-\\rence}}{Setup}} & \shortstack{\textsc{time-}\\\textsc{stratified}}  & \shortstack{\textsc{time-}\\\textsc{stratified}\\ + \\dynamic eval} \\\hline
\multicolumn{2}{c|}{All words}  & 22.45& 22.17\\\hline
\multicolumn{2}{c|}{All \LFTRHFTE}  & 109.73& 66.26\\\hline
 \multicolumn{2}{c|}{1st} & 694.95 & 357.40\\\hline
\multirow{2}{*}{2nd} &  \shortstack{1st in \\ memory} & 75.95& 44.21\\\cdashline{2-4}\\[-0.7em]
     & \shortstack{1st \textbf{NOT} \\ in memory} & 2,719.25 & 1,430.34\\\bottomrule
\end{tabular}}
\caption{Perplexity of \timestratified~model on \LFTRHFTE~on \wmt, broken down by whether the word is encountered for the 1st or the 2nd time in the test document, and for the latter, whether the 1st occurrence was in the TXL context. The last column shows results with dynamic evaluation (\S\ref{sec:solution}).}
\label{tab:adaptation_speed}
\vspace{-10pt}
\end{wraptable}

Concretely, we define \LFTRHFTE~as those that occur frequently on the test set (at least 50 times), but either:
(i) were previously unseen on the training set, or
(ii) occurred much less frequently on the training set than on the test set, as indicated by an at least 5 times lower unigram probability. This procedure yields a reasonably-sized set of 287 \LFTRHFTE~and 87,636 mentions in our 2018-2019 test documents. 
Many of these words indeed reflect strong temporal dynamics: e.g. ``Ardern'' (who became the New Zealand PM in late-2017) and ``Novichok'' (which is what Sergey and Yulia Skripal were poisoned with in 2018). Fig.~\ref{tab:adaptation_speed} shows that the \timestratified~model performs substantially worse for \LFTRHFTE---an almost 5x worse perplexity (110 vs 22) than the overall one (Figure~\ref{tab:overall_perplexity}).

\vspace{-4mm}
\paragraph{Perplexity of first and second occurrences of \LFTRHFTE.} We now ask: How well can Transformer LMs rapidly adapt to new information and \LFTRHFTE? Concretely, LMs that perform well in our non-stationary world should be able to predict subsequent occurrences of \LFTRHFTE~(e.g. ``COVID-19''), which exhibit strong temporal dynamics, much better than the first occurrences of these words, because these words appear frequently on the test set prefix---even though these \LFTRHFTE~do not appear as frequently on the training set. In Fig.~\ref{tab:adaptation_speed}, we show the  perplexity obtained by the \timestratified~model under two conditions: For the first and second occurrences of \LFTRHFTE~in a test document.                                                                                                                                                                               

Although the model has a high ppl. the first time it generates \LFTRHFTE~in the document (ppl. of $\sim$694.95), it has a much lower ppl. for generating the same words for the second time, but \emph{only if} the first word is available in the Transformer context. In such case, the model can simply copy the same word from the context; this finding reaffirms the strong copying ability of the attention block~\citep{bahdanau_15,vinyals_2015}. This means that the ability of Transformers to condition on long-range context is \emph{already} a useful feature for temporal generalization, even when we are not explicitly updating the model parameters with new data. However, we observe no such effect when the first occurrence falls \emph{outside} of the Transformer memory (ppl. of >2,700), highlighting the need to scale Transformers to even longer sequences \citep[\emph{inter alia}]{child_2019,correia-etal-2019-adaptively,kitaev_2020,beltagy2020longformer} to improve temporal generalization.

\vspace{-4mm}
\paragraph{Importance.} Our analyses provide a targeted evaluation of temporal generalization in LMs, which enable us to benchmark progress precisely on things that matter the most for temporal generalization (e.g. evaluating LMs on named entities, fast-changing topics, and adaptation speed to \LFTRHFTE, rather than relying on overall ppl. as a sole metric for measuring LM progress).

\vspace{-2mm}
\section{The effect of outdated models persists even when increasing model sizes}
\label{sec:model_size}

\begin{wrapfigure}{R}{0.5\textwidth}
\vspace{-10pt}
\centering
\includegraphics[width=9cm,height=5cm,keepaspectratio]{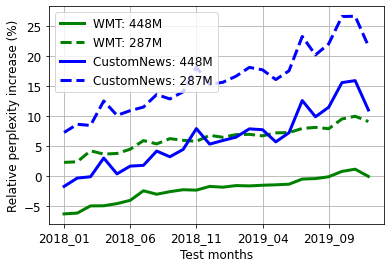}
\caption{Relative perplexity increase of the \timestratified~models with 287M (dotted line) and 448M parameters (solid line), respectively, over the \control~model with 287M parameters, for \wmt~and \customnews~(\S\ref{sec:model_size}).}
\label{fig:model_size}
\vspace{-20pt}
\end{wrapfigure}
Recently, increasing model size has led to substantial improvements in perplexity, downstream tasks, and few-shot learning ability~\citep{Kaplan:etal:2020,Brown:etal:2020}. But can increasing model size also improve temporal generalization? To this end,  we train a bigger \timestratified~model with 448M parameters---a 60\% increase over the previous 287M model and 30\% larger than GPT-2$_{\text{MEDIUM}}$.

Similar to Section~\ref{sec:overall_ppl}, we report the respective perplexity increase of the newly trained \timestratified$^{448M}$ model over the \control$^{287M}$ model (solid lines). We reproduce the relative perplexity increase of the smaller \timestratified$^{287M}$ model over the \control$^{287M}$ one (Fig.~\ref{fig:overtime_pos}) as the dotted lines. 

If increasing the model size was able to delay temporal degradation, we would expect to see the solid lines produced by the bigger models to have reduced (i.e., flatter) slopes compared to the dotted lines produced by the smaller models. While larger \timestratified~models, as expected, achieve lower absolute perplexities (5.5\% improvement), model size has \emph{no significant effect} on the slope of these lines ($p>0.05$, assessed using a t-test on the slopes found by fitting a linear regression). On both datasets, by the end of the test period (i.e. late-2019), a smaller but more up-to-date \control$^{287M}$ model outperforms a 60\% larger but two-year out-of-date \timestratified$^{448M}$ model. Hence, building models that perform well in this setup 
requires solutions that more directly tackle the specific challenges we emphasized through our findings so far, and update the model's knowledge with new information.

\vspace{-2mm}
\section{Time-stratified question answering}
\label{sec:downstream}
So far we have evaluated the LMs intrinsically, through perplexity, which is important because language modelling is a \emph{foundational} task that affects many NLP applications through language model pretraining. However, we still do not know how this perplexity deterioration affects practical applications of LMs, i.e., how do out-of-date LMs affect different types of downstream tasks?

\begin{wrapfigure}{R}{0.5\textwidth}
    \centering
    \includegraphics[width=9cm,height=5cm,keepaspectratio]{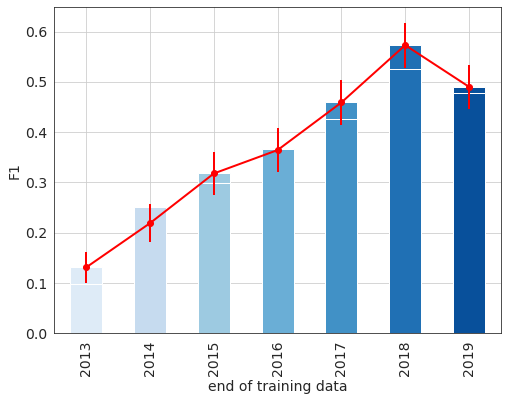}
\caption{Synthetic questions about political figures: Model performance as we shift the end of the training set chronologically away from the year about which we ask questions (i.e, 2019). The error bars indicate two standard errors of mean.}
\label{fig:synth_qa}
\end{wrapfigure}
\vspace{-2mm}
\paragraph{Closed-book question answering (QA).}
Closed-book QA is a popular testbed for evaluating pretrained LMs that have to compress and encode knowledge found in a big corpus. 
But given the relative lack of existing time-stamped, news QA datasets that evaluate LMs' ability to answer questions about events that happen outside of their training period in a closed-book fashion, we construct a dataset of synthetic questions the government officials using the following template: ``\emph{Who is the [government role] of [country/state] in [month/year]?}'' In total, we construct a test set of 438 questions about 22 different government roles from 11 countries (see Appendix~\ref{sec:appendix_downstream} for examples). We pretrain the TXL model (as described in Section \ref{sec:model}) using the \wmt\ dataset up to the years 2012, 2013,\dots, 2019, respectively. We then fine-tune all these models to answer questions about government officials for 2011 to get the model accustomed to the task format, and evaluate on synthetic questions related to the year 2019. Fig.~\ref{fig:synth_qa} shows the substantial accuracy deterioration as we shift the end of the pretraining data away from 2019, the year for which we ask questions. This finding demonstrates how the fine-tuned LMs' lack of more recent factual knowledge affects their performance on this task. Note that the slight drop in accuracy in 2019 compared to 2018 is due to dataset noise. Anecdotally, we observe that the 2019 model mixes up the names of Russian and American presidents, which often co-occurred in the same context in 2019.

\vspace{-4mm}
\paragraph{Reading comprehension.} Nevertheless, we do not expect all downstream tasks to be equally affected by outdated LMs. To illustrate this point, we perform a reading comprehension experiment using NewsQA \citep{trischler-etal-2017-newsqa}, where the evidence documents are presented together with the questions into the prefix of the model. Hence, the model has all necessary information to answer the questions, and thus outdated LMs will likely present less of a challenge in this type of tasks. 
We obtain a \timestratified~NewsQA by recovering the articles' timestamps.\footnote{https://cs.nyu.edu/~kcho/DMQA/}. We test on questions from 2009, for a of total 10000 questions (see Appendix~\ref{sec:appendix_downstream} for question-answer examples). We evaluate how well LMs trained on \customnews~until the end of 2008 performs in comparison to LMs trained until the end of 2009: Both models perform identically at 0.47 F1. Hence, time-stratified evaluations for reading comprehension, where the answers are extractive and can be copied from the passage, pose less of a challenge for outdated LMs, unlike knowledge-intensive, closed-book QA.
\vspace{-2mm}
\section{Keeping models up-to-date: Online learning through dynamic evaluation}
\label{sec:solution}
One way to mitigate LMs' degradation over time is to continually update the models' knowledge with new information as new documents arrive into the stream. One of the ways to do this is through dynamic evaluation \citep{mikolov_etal_2010,graves_2013,Krause:etal:2019}---a form of online learning that continually updates the parameters of a pretrained model by performing gradient descent on the new data. While most prior work used dynamic evaluation to perform updates within a document, hence adapting to local topical shifts, here we use it to adapt to the temporal dynamics that occur within a stream of chronologically ordered documents, hence adapting to temporal trends across documents. Appendix \ref{sec:appendix_dynamic} has more details on dynamic evaluation and our empirical settings.
\begin{wrapfigure}{R}{.5\textwidth}
\begin{center}
\includegraphics[width=9cm,height=5cm,keepaspectratio]{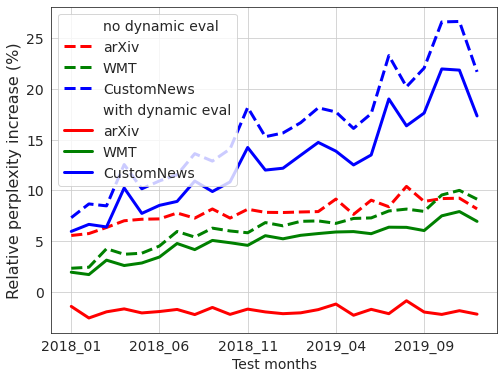}
\caption{Relative perplexity increase with (solid lines) and without (dotted lines) dynamic evaluation, for the \timestratified~model.\label{fig:dynamic_eval}}
\end{center}
\end{wrapfigure}

We plot the results in Fig.~\ref{fig:dynamic_eval}: Dotted lines reflect the perplexity increase when comparing the \control~model to the \timestratified~model, i.e., the same graph as in Fig.~\ref{fig:over_month_all}, whereas solid lines reflect the perplexity increase achieved when comparing the same \control~model with the \timestratified~model augmented with dynamic evaluation (\timestratified$^{dyn}$). In all datasets, dynamic evaluation reduces the speed of the model becoming outdated, as evidenced by the \emph{reduced} upward slope, with a significant effect for \arxiv~and \wmt~ ($p<0.05$, assessed using a t-test on the slopes found by fitting a linear regression). The improvements are  more pronounced for \arxiv, where a more granular analysis over weeks reveals that the model needs only about 
one week worth of data to overtake the \control~model. 
Moreover, we see much larger improvements for predicting \LFTRHFTE, which exhibit strong temporal dynamics (\S\ref{sec:adaptation}, see Fig.~\ref{tab:adaptation_speed}): We observe a 39.62\% ppl. reduction from 109.73 to 66.2 for \LFTRHFTE, compared to the overall ppl. reduction (a 1.25\% reduction from 22.45 to 22.17 for \wmt; Fig.~\ref{tab:dynamic_eval_params}).
\begin{wraptable}{R}{.5\textwidth}
\resizebox{0.48\textwidth}{!}{%
\begin{tabular}{l|c c c}\toprule
    \shortstack[l]{Parameters\\that get updated }& \wmt & \shortstack[m]{\textbf{\textsc{Custom}}\\\textbf{\textsc{News}}} & \arxiv \\\hline
    All parameters  &22.17 & \textbf{20.72} & \textbf{20.98} \\
    Bias-only      & \textbf{22.16} &20.96 & 21.24 \\
    Embeddings-only &22.32 & 21.21 & 22.27\\
    \cline{1-4}
     no dynamic eval. &22.45 &21.33 & 23.07 \\
     \bottomrule
\end{tabular}}
\caption{Perplexity of \timestratified~model when updating some of the parameters.\label{tab:dynamic_eval_params}}
\vspace{-10pt}
\end{wraptable}

When aiming to keep models up-to-date (especially for larger models), lightweight yet effective approaches are preferable because they allow the model to rapidly digest new information with minimal time, computation, and carbon costs. 
We thus experiment with updating only the embedding layer (i.e., 52M parameters), capturing lexical semantic changes, as well as updating only the bias terms at all layers (i.e., 198K parameters), as recently introduced by~\citet{BenZake:etal:2021}. Fig.~\ref{tab:dynamic_eval_params} presents the results: In line with the findings of~\citet{BenZake:etal:2021}, updating only the bias terms performs nearly as well as updating the full model. 

\vspace{-2mm}
\paragraph{Beyond dynamic evaluation} We remark that dynamic evaluation alone, while effective, does not fully solve temporal generalization, as evidenced by the prevailing (albeit gentler) upward slopes on \wmt~and \customnews. We expect that larger gains can be achieved by fully embracing continual learning in LMs---striking a balance between quickly adapting to new data and achieving positive forward transfer \citep{GradientEpisodicMemory}, while avoiding catastrophic forgetting \citep{mccloskey:catastrophic,Kirkpatrick3521}. Indeed, as we show in Figure~\ref{fig:catastrophic_forgetting}, while dynamic evaluation is able to improve generalization to future data, it causes catastrophic forgetting of the past data. Furthermore, recent semi-parametric models \citep{Guu:etal:2020,rag,dpr,knn-lm,Yogatama:etal:2021} lend themselves well to continual learning, where new knowledge can be stored in an external memory, which can be updated without retraining the whole model. A related approach is to disentangle the acquisition of up-to-date knowledge from the language learning itself, and enable direct editing of that knowledge within the model~\citep{Sinitsin:etal:2020,Zhu:etal:2020,DeCao:etal:2021}.

\vspace{-2mm}
\section{Related Work}
\label{sec:background}

\paragraph{Concept drift.} Detecting changes in data streams, also known as concept drift, has a long history in machine learning \citep{concept_drift_widmer_1996,kifer2004detecting,baena2006early,dries2009adaptive,concept_drift_review_2019}. In NLP, most recent work in this area models lexical change by training word embeddings \citep{Hamilton2016diachronic,Szymanski2017analogies,yin2018global} and deep neural networks \citep{rosenfeld2018deep,bjerva2019back} on data of different time spans.

\vspace{-4mm}
\paragraph{Out-of-Distribution (OoD) generalization.} OoD generalization is well-studied in NLP \citep{Blitzer:etal:2006,Daume:etal:2007,Axelrod:etal:2011}, and has recently been addressed for neural LMs and transfer learning \citep{Fried:etal:2019,Oren:etal:2019,Gururangan:etal:2020}, where pretrained LMs lead to substantial improvements and increased robustness \citep{Hendrycks:etal:2020}. 
While most prior work has focused on distributional shifts in terms of topic and domain~\citep{Kruszewski:etal:2020}, distributional shifts in terms of \emph{time} also constitute an important and realistic challenge that NLP systems of today (including those based on large LM pretraining) must be able to perform well at.

\vspace{-4mm}
\paragraph{Continual learning \& streaming LMs.} Our work is closely related to continual and lifelong learning, which aim to design models that continually accumulate new knowledge without forgetting relevant information about the past \citep{mccloskey:catastrophic,THRUN199525,FRENCH1999128,mitchell_2015,rusu_etal_2016,Kirkpatrick3521,shedivat_etal_2018,HADSELL20201028}. The distribution of words and context in natural language changes rapidly with time, and hence constitutes an important test bed for developing continual learning systems.
More specific to the LM literature, prior work proposed ways of designing LMs that efficiently adapt their knowledge to continuous streams of new information \citep[\emph{inter alia}]{jelinek_1991,Wang:etal:2008,goyal-2009,Osborne:VanDurme:2014,Yogatama:etal:2014}---often known as \emph{streaming LMs}, albeit mostly in the context of $n$-gram LMs. While we show that Transformer LMs achieve much better perplexity than previous $n$-gram models (the Transformer LM ppl. in \S\ref{sec:experiment1} are substantially better than those in prior streaming $n$-gram LM literature), we show that Transformer LMs similarly suffer from the temporal degradation problem. Given the different nature of our LMs today (i.e. deep neural models rather than $n$-gram LMs), we argue that now is the right time to make progress on this open research question, with notable progress in other NLP tasks \citep{Masson2019episodic,Sun2020lamal}.

\vspace{-4mm}
\paragraph{Temporal splits in NLP.} Prior work has used temporal splits (i.e. training on text from the past and evaluating on future text) for NLP tasks like machine translation \citep{levenberg_2010}, sentiment analysis \citep{lukes-sogaard-2018-sentiment}, named entity recognition \citep{fromreide-etal-2014-crowdsourcing,rijhwani-preotiuc-pietro-2020-temporally}, and others \citep{searchqa,bjerva_2020,Sogaard:etal:2020}. Nevertheless, the vast majority of NLP benchmarks today still do not perform temporal splits, and hence do not measure how well models can generalize to future data. Furthermore, this work has two key distinctions from prior work. First, we focus on language modelling---a \emph{foundational} task that is used for many NLP systems today through LM pretraining---and propose a benchmark to measure progress in this direction. Second, we go one step further than prior work and perform a thorough analysis to pinpoint \emph{what kinds} of predictions the model is struggling with. Such analysis can then be used to better measure progress in dynamic language modelling, where improvements are not always easily discernible from overall ppl. alone (e.g. performance on \LFTRHFTE; \S\ref{sec:analysis}).

\vspace{-4mm}
\section{Conclusion}
\label{sec:discussion}
We evaluated the extent to which our current language models can generalize well in the realistic setup of predicting future utterances outside their training period. We found that current practices that train and evaluate models on data from overlapping time period \emph{overestimate} model generalization to future utterances, and that Transformer LMs become increasingly outdated with time. We found that increasing model size alone---a key driver behind recent LM success---is not a solution for this problem, and that this degradation affects downstream tasks that require up-to-date factual knowledge. 

\vspace{-4mm}
\paragraph{Generality to other domains.} The importance of temporal generalization extends beyond language modelling and NLP. Many other commercial machine learning systems like speech processing and video understanding also involve non-stationary data, and are similarly trained on data that were collected in the past, but deployed on new data from the future. Since many of these tasks also use Transformers \citep[\emph{inter alia}]{girdhar2019video,gulati2020conformer}, we expect our findings to generalize to these domains, although a more complete investigation to this end is left to future work.
\vspace{-4mm}
\paragraph{Limitations} While we explored how the LM performance degradation with time affects two types of question-answering tasks, a broader variety of tasks is needed to obtain a more holistic  picture on how temporal generalization manifests in downstream tasks. An open research question is thus how we can create and maintain benchmarks that are not static~\citep{Nie:etal:2019, Potts:etal:2020} and further promote research on continual and life-long learning. 

\vspace{-4mm}
\paragraph{Is this all obvious?} Our findings should not come us a surprise: That the world around us changes with time---and thus \emph{what} and \emph{how} we talk about it also evolve accordingly---is hardly controversial. But for the most part, these temporal dynamics are \emph{still not} currently reflected in the way that we train and evaluate our neural LMs. Our aim here is to highlight how such static evaluations overestimate models' performance, especially around predictions that relate to factual knowledge, which constitute an important use case of NLP systems today. With the compilation of ever-larger web crawls for LM pretraining~\citep{Gao:etal:2021}, now is the right time to rethink how our splits are constructed~\citep{Sogaard:etal:2020}, construct temporal splits that evaluate models on their ability to generalize to future data, and include timestamp information in both pretraining datasets and downstream tasks to facilitate this kind of more realistic LM evaluation. This strategy will not only allow us to assess models' performance on a realistic type of out-of-sample data, but also circumvent test data contamination affecting both LM and downstream task evaluations more broadly, e.g., in widely used QA datasets like Natural Questions \citep{natural-questions} and TriviaQA \citep{joshi-etal-2017-triviaqa}, which have been shown to contain alarmingly high proportions of overlapping training and test data~\citep{Lewis:etal:2020}. Our dynamic, streaming LM benchmarks---alongside the evaluation metrics that evaluate LMs on things that matter the most for temporal generalization (e.g. named entities, \LFTRHFTE)---will be released to encourage more research in this area, and reward the development of \textbf{adaptive} language models that can remain up-to-date with respect to our non-stationary world.    
\section{Broader Societal Impact Discussion}
\label{sec:societal_impact}
Lastly, we remark on two aspects of the broader societal impact pertaining to the importance of continually-updating language models. First, we argue that having NLP models---the vast majority of which are built on top of pretrained language models---that can remain up-to-date with our current social trends and public perception is relevant for mitigating potential harms and biases caused by NLP models. For instance, recently there has been renewed public support and interest for social justice movements in 2020, such as the \#BlackLivesMatter movement \citep{blm_support}. Hence, without explicit mechanisms to update the models’ knowledge, language models that were trained before this time period can potentially miss out on shifting language used to describe such movements---where such movements are now more widely supported by the general public---and potentially produce outdated, biased language that is no longer frequently used at present. On the other hand, we should also be careful not to let the model update its knowledge with material that can add or amplify to the bias and prejudice of the model \citep{chang-etal-2019-bias,Bender:etal:2021}.

Second, our findings highlight the risk of the “brute-force” approach of keeping models up-to-date by periodically retraining the model from scratch, for instance by combining the old and new data. Given the increasing size of NLP models, training large models from scratch each time incurs increasingly more expensive computational and environmental costs \citep{strubell-2019,carbon-impact-2021}. Hence, our findings emphasise the need for more efficient and lightweight approaches of keeping models up-to-date, whilst mitigating catastrophic forgetting at the same time. Our work provides a benchmark to measure progress in this space, and we strongly encourage future work that uses our benchmark to also report the computational costs of their approach for keeping language models up-to-date. Lastly, we remark that the ethical considerations and risks of working with large language models also apply to our work \citep{Bender:etal:2021}.

\begin{ack}
We thank Paul Michel, Laura Rimell, and Chris Dyer for  useful feedback throughout the different stages of this project. We would also like to thank Katie Millican, Sebastian Borgeaud, Trevor Cai, Roman Ring, Jack Rae, and Geoffrey Irving for their initial work on the codebase.
\end{ack}

\bibliographystyle{plainnat}
\bibliography{references}

\begin{thebibliography}{92}
\providecommand{\natexlab}[1]{#1}
\providecommand{\url}[1]{\texttt{#1}}
\expandafter\ifx\csname urlstyle\endcsname\relax
  \providecommand{\doi}[1]{doi: #1}\else
  \providecommand{\doi}{doi: \begingroup \urlstyle{rm}\Url}\fi

\bibitem[Al{-}Shedivat et~al.(2018)Al{-}Shedivat, Bansal, Burda, Sutskever,
  Mordatch, and Abbeel]{shedivat_etal_2018}
Maruan Al{-}Shedivat, Trapit Bansal, Yuri Burda, Ilya Sutskever, Igor Mordatch,
  and Pieter Abbeel.
\newblock Continuous adaptation via meta-learning in nonstationary and
  competitive environments.
\newblock In \emph{Proc. of ICLR}, 2018.

\bibitem[Augenstein et~al.(2019)Augenstein, Lioma, Wang, Chaves~Lima, Hansen,
  Hansen, and Simonsen]{Augenstein2019}
Isabelle Augenstein, Christina Lioma, Dongsheng Wang, Lucas Chaves~Lima, Casper
  Hansen, Christian Hansen, and Jakob~Grue Simonsen.
\newblock {M}ulti{FC}: A real-world multi-domain dataset for evidence-based
  fact checking of claims.
\newblock In \emph{Proc. of EMNLP-IJCNLP}, 2019.

\bibitem[Axelrod et~al.(2011)Axelrod, He, and Gao]{Axelrod:etal:2011}
Amittai Axelrod, Xiaodong He, and Jianfeng Gao.
\newblock Domain adaptation via pseudo in-domain data selection.
\newblock In \emph{Proc. of EMNLP}, 2011.

\bibitem[Baena-Garc{\i}a et~al.(2006)Baena-Garc{\i}a, del Campo-{\'A}vila,
  Fidalgo, Bifet, Gavalda, and Morales-Bueno]{baena2006early}
Manuel Baena-Garc{\i}a, Jos{\'e} del Campo-{\'A}vila, Ra{\'u}l Fidalgo, Albert
  Bifet, R~Gavalda, and R~Morales-Bueno.
\newblock Early drift detection method.
\newblock In \emph{Fourth international workshop on knowledge discovery from
  data streams}, volume~6, pages 77--86, 2006.

\bibitem[Bahdanau et~al.(2015)Bahdanau, Cho, and Bengio]{bahdanau_15}
Dzmitry Bahdanau, Kyunghyun Cho, and Yoshua Bengio.
\newblock Neural machine translation by jointly learning to align and
  translate.
\newblock In \emph{Proc. of ICLR}, 2015.

\bibitem[Beltagy et~al.(2020)Beltagy, Peters, and Cohan]{beltagy2020longformer}
Iz~Beltagy, Matthew~E. Peters, and Arman Cohan.
\newblock Longformer: The long-document transformer.
\newblock \emph{CoRR}, abs/2004.05150, 2020.

\bibitem[Ben-Zaken et~al.(2021)Ben-Zaken, Ravfogel, and
  Goldberg]{BenZake:etal:2021}
Elad Ben-Zaken, Shauli Ravfogel, and Yoav Goldberg.
\newblock Bitfit: Simple parameter-efficient fine-tuning for transformer-based
  masked language-models, 2021.

\bibitem[Bender et~al.(2021)Bender, Gebru, McMillan-Major, and
  Shmitchell]{Bender:etal:2021}
Emily~M. Bender, Timnit Gebru, Angelina McMillan-Major, and Shmargaret
  Shmitchell.
\newblock On the dangers of stochastic parrots: Can language models be too big?
\newblock In \emph{Proceedings of FAccT 2021}, 2021.

\bibitem[Bjerva et~al.(2019)Bjerva, Kouw, and Augenstein]{bjerva2019back}
Johannes Bjerva, Wouter~M Kouw, and Isabelle Augenstein.
\newblock Back to the future--sequential alignment of text representations.
\newblock In \emph{34rd AAAI Conference on Artificial Intelligence}, pages
  1909--03464. Association for the Advancement of Artificial Intelligence,
  2019.

\bibitem[Bjerva et~al.(2020)Bjerva, Kouw, and Augenstein]{bjerva_2020}
Johannes Bjerva, Wouter Kouw, and Isabelle Augenstein.
\newblock Back to the future - temporal adaptation of text representations.
\newblock In \emph{Proc. of AAAI}, 2020.

\bibitem[Blei et~al.(2003)Blei, Ng, and Jordan]{Blei_lda}
David~M. Blei, Andrew~Y. Ng, and Michael.~I Jordan.
\newblock Latent dirichlet allocation.
\newblock \emph{Journal of Machine Learning Research}, 3, 2003.

\bibitem[Blitzer et~al.(2006)Blitzer, McDonald, and Pereira]{Blitzer:etal:2006}
John Blitzer, Ryan McDonald, and Fernando Pereira.
\newblock Domain adaptation with structural correspondence learning.
\newblock In \emph{Proc. of EMNLP}, 2006.

\bibitem[Brown et~al.(2020)Brown, Mann, Ryder, Subbiah, Kaplan, Dhariwal,
  Neelakantan, Shyam, Sastry, Askell, Agarwal, Herbert-Voss, Krueger, Henighan,
  Child, Ramesh, Ziegler, Wu, Winter, Hesse, Chen, Sigler, Litwin, Gray, Chess,
  Clark, Berner, McCandlish, Radford, Sutskever, and Amodei]{Brown:etal:2020}
Tom~B. Brown, Benjamin Mann, Nick Ryder, Melanie Subbiah, Jared Kaplan,
  Prafulla Dhariwal, Arvind Neelakantan, Pranav Shyam, Girish Sastry, Amanda
  Askell, Sandhini Agarwal, Ariel Herbert-Voss, Gretchen Krueger, Tom Henighan,
  Rewon Child, Aditya Ramesh, Daniel~M. Ziegler, Jeffrey Wu, Clemens Winter,
  Christopher Hesse, Mark Chen, Eric Sigler, Mateusz Litwin, Scott Gray,
  Benjamin Chess, Jack Clark, Christopher Berner, Sam McCandlish, Alec Radford,
  Ilya Sutskever, and Dario Amodei.
\newblock Language models are few-shot learners.
\newblock \emph{CoRR}, abs/2005.14165, 2020.

\bibitem[Chang et~al.(2019)Chang, Prabhakaran, and
  Ordonez]{chang-etal-2019-bias}
Kai-Wei Chang, Vinod Prabhakaran, and Vicente Ordonez.
\newblock Bias and fairness in natural language processing.
\newblock In \emph{Proc. of EMNLP-IJCNLP: Tutorial Abstracts}, 2019.

\bibitem[Chelba et~al.(2013)Chelba, Mikolov, Schuster, Ge, Brants, Koehn, and
  Robinson]{Chelba:etal:2013}
Ciprian Chelba, Tomas Mikolov, Mike Schuster, Qi~Ge, Thorsten Brants, Phillipp
  Koehn, and Tony Robinson.
\newblock One billion word benchmark for measuring progress in statistical
  language modeling.
\newblock \emph{arXiv preprint arXiv:1312.3005}, 2013.

\bibitem[Child et~al.(2019)Child, Gray, Radford, and Sutskever]{child_2019}
Rewon Child, Scott Gray, Alec Radford, and Ilya Sutskever.
\newblock Generating long sequences with sparse transformers.
\newblock \emph{CoRR}, abs/1904.10509, 2019.

\bibitem[Cohn and Quealy(2020)]{blm_support}
Nate Cohn and Kevin Quealy.
\newblock How public opinion has moved on black lives matter.
\newblock \emph{The New York Times}, 2020.
\newblock URL
  \url{https://www.nytimes.com/interactive/2020/06/10/upshot/black-lives-matter-attitudes.html}.

\bibitem[Correia et~al.(2019)Correia, Niculae, and
  Martins]{correia-etal-2019-adaptively}
Gon{\c{c}}alo~M. Correia, Vlad Niculae, and Andr{\'e} F.~T. Martins.
\newblock Adaptively sparse transformers.
\newblock In \emph{Proc. of EMNLP-IJCNLP}, 2019.

\bibitem[Dai et~al.(2019)Dai, Yang, Yang, Carbonell, Le, and
  Salakhutdinov]{Dai:etal:2019}
Zihang Dai, Zhilin Yang, Yiming Yang, Jaime Carbonell, Quoc~V. Le, and Ruslan
  Salakhutdinov.
\newblock Transformer-xl: Attentive language models beyond a fixed-length
  context.
\newblock In \emph{Proc. of ACL}, 2019.

\bibitem[Daum{\'e}~III(2007)]{Daume:etal:2007}
Hal Daum{\'e}~III.
\newblock Frustratingly easy domain adaptation.
\newblock In \emph{Proc. of ACL}, 2007.

\bibitem[d'Autume et~al.(2019)d'Autume, Ruder, Kong, and
  Yogatama]{Masson2019episodic}
Cyprien de~Masson d'Autume, Sebastian Ruder, Lingpeng Kong, and Dani Yogatama.
\newblock {Episodic Memory in Lifelong Language Learning}.
\newblock In \emph{Proc. of NeurIPS}, 2019.

\bibitem[De~Cao et~al.(2021)De~Cao, Aziz, and Titov]{DeCao:etal:2021}
Nicola De~Cao, Wilker Aziz, and Ivan Titov.
\newblock Editing factual knowledge in language models.
\newblock \emph{arXiv preprint arXiv:2104.08164}, 2021.

\bibitem[Ding et~al.(2015)Ding, Zhang, Liu, and Duan]{ding_2015}
Xiao Ding, Yue Zhang, Ting Liu, and Junwen Duan.
\newblock Deep learning for event-driven stock prediction.
\newblock In \emph{Proc. of IJCAI}, 2015.

\bibitem[Dries and R{\"u}ckert(2009)]{dries2009adaptive}
Anton Dries and Ulrich R{\"u}ckert.
\newblock Adaptive concept drift detection.
\newblock \emph{Statistical Analysis and Data Mining: The ASA Data Science
  Journal}, 2\penalty0 (5-6):\penalty0 311--327, 2009.

\bibitem[Dunn et~al.(2017)Dunn, Sagun, Higgins, G{\"u}ney, Cirik, and
  Cho]{searchqa}
Matthew Dunn, Levent Sagun, Mike Higgins, V.~U. G{\"u}ney, Volkan Cirik, and
  Kyunghyun Cho.
\newblock Searchqa: A new q\&a dataset augmented with context from a search
  engine.
\newblock \emph{ArXiv}, abs/1704.05179, 2017.

\bibitem[Fedus et~al.(2021)Fedus, Zoph, and Shazeer]{Fedus:etal:2021}
William Fedus, Barret Zoph, and Noam Shazeer.
\newblock Switch transformers: Scaling to trillion parameter models with simple
  and efficient sparsity.
\newblock \emph{arXiv preprint arXiv:2101.03961}, 2021.

\bibitem[French(1999)]{FRENCH1999128}
Robert~M. French.
\newblock Catastrophic forgetting in connectionist networks.
\newblock \emph{Trends in Cognitive Sciences}, 3\penalty0 (4), 1999.

\bibitem[Fried et~al.(2019)Fried, Kitaev, and Klein]{Fried:etal:2019}
Daniel Fried, Nikita Kitaev, and Dan Klein.
\newblock Cross-domain generalization of neural constituency parsers.
\newblock In \emph{Proc. of ACL}, 2019.

\bibitem[Fromreide et~al.(2014)Fromreide, Hovy, and
  S{\o}gaard]{fromreide-etal-2014-crowdsourcing}
Hege Fromreide, Dirk Hovy, and Anders S{\o}gaard.
\newblock Crowdsourcing and annotating {NER} for {T}witter {\#}drift.
\newblock In \emph{Proc. of LREC}, 2014.

\bibitem[Gao et~al.(2021)Gao, Biderman, Black, Golding, Hoppe, Foster, Phang,
  He, Thite, Nabeshima, Presser, and Leahy]{Gao:etal:2021}
Leo Gao, Stella Biderman, Sid Black, Laurence Golding, Travis Hoppe, Charles
  Foster, Jason Phang, Horace He, Anish Thite, Noa Nabeshima, Shawn Presser,
  and Connor Leahy.
\newblock The pile: An 800gb dataset of diverse text for language modeling,
  2021.

\bibitem[Girdhar et~al.(2019)Girdhar, Carreira, Doersch, and
  Zisserman]{girdhar2019video}
Rohit Girdhar, Jo\~{a}o Carreira, Carl Doersch, and Andrew Zisserman.
\newblock {Video Action Transformer Network}.
\newblock In \emph{Proc. of CVPR}, 2019.

\bibitem[Goyal et~al.(2009)Goyal, Daum{\'e}~III, and
  Venkatasubramanian]{goyal-2009}
Amit Goyal, Hal Daum{\'e}~III, and Suresh Venkatasubramanian.
\newblock Streaming for large scale {NLP}: Language modeling.
\newblock In \emph{Proc. of NAACL-HLT}, 2009.

\bibitem[Graves(2013)]{graves_2013}
Alex Graves.
\newblock Generating sequences with recurrent neural networks.
\newblock \emph{CoRR}, abs/1308.0850, 2013.

\bibitem[Gulati et~al.(2020)Gulati, Qin, Chiu, Parmar, Zhang, Yu, Han, Wang,
  Zhang, Wu, and Pang]{gulati2020conformer}
Anmol Gulati, James Qin, Chung-Cheng Chiu, Niki Parmar, Yu~Zhang, Jiahui Yu,
  Wei Han, Shibo Wang, Zhengdong Zhang, Yonghui Wu, and Ruoming Pang.
\newblock Conformer: Convolution-augmented transformer for speech recognition.
\newblock In \emph{Proc. of INTERSPEECH}, 2020.

\bibitem[Gururangan et~al.(2020)Gururangan, Marasovi{\'c}, Swayamdipta, Lo,
  Beltagy, Downey, and Smith]{Gururangan:etal:2020}
Suchin Gururangan, Ana Marasovi{\'c}, Swabha Swayamdipta, Kyle Lo, Iz~Beltagy,
  Doug Downey, and Noah~A. Smith.
\newblock Don{'}t stop pretraining: Adapt language models to domains and tasks.
\newblock In \emph{Proc. of ACL}, 2020.

\bibitem[Guu et~al.(2020)Guu, Lee, Tung, Pasupat, and Chang]{Guu:etal:2020}
Kelvin Guu, Kenton Lee, Zora Tung, Panupong Pasupat, and Ming-Wei Chang.
\newblock Retrieval-augmented language model pre-training.
\newblock In \emph{Proc. of ICML}, 2020.

\bibitem[Hadsell et~al.(2020)Hadsell, Rao, Rusu, and Pascanu]{HADSELL20201028}
Raia Hadsell, Dushyant Rao, Andrei~A. Rusu, and Razvan Pascanu.
\newblock Embracing change: Continual learning in deep neural networks.
\newblock \emph{Trends in Cognitive Sciences}, 24\penalty0 (12), 2020.

\bibitem[Hamilton et~al.(2016)Hamilton, Leskovec, and
  Jurafsky]{Hamilton2016diachronic}
William~L Hamilton, Jure Leskovec, and Dan Jurafsky.
\newblock {Diachronic Word Embeddings Reveal Statistical Laws of Semantic
  Change}.
\newblock In \emph{Proc. of ACL}, pages 1489--1501, 2016.

\bibitem[Hendrycks et~al.(2020)Hendrycks, Liu, Wallace, Dziedzic, Krishnan, and
  Song]{Hendrycks:etal:2020}
Dan Hendrycks, Xiaoyuan Liu, Eric Wallace, Adam Dziedzic, Rishabh Krishnan, and
  Dawn Song.
\newblock Pretrained transformers improve out-of-distribution robustness.
\newblock In \emph{Proc. of ACL}, 2020.

\bibitem[Jelinek et~al.(1991)Jelinek, Merialdo, Roukos, and
  Strauss]{jelinek_1991}
F.~Jelinek, B.~Merialdo, S.~Roukos, and M.~Strauss.
\newblock A dynamic language model for speech recognition.
\newblock In \emph{Speech and Natural Language: Proceedings of a Workshop Held
  at Pacific Grove, California, {F}ebruary 19-22, 1991}, 1991.

\bibitem[Joshi et~al.(2017{\natexlab{a}})Joshi, Choi, Weld, and
  Zettlemoyer]{joshi-etal-2017-triviaqa}
Mandar Joshi, Eunsol Choi, Daniel Weld, and Luke Zettlemoyer.
\newblock {T}rivia{QA}: A large scale distantly supervised challenge dataset
  for reading comprehension.
\newblock In \emph{Proc. of ACL}, 2017{\natexlab{a}}.

\bibitem[Joshi et~al.(2017{\natexlab{b}})Joshi, Choi, Weld, and
  Zettlemoyer]{triviaqa}
Mandar Joshi, Eunsol Choi, Daniel Weld, and Luke Zettlemoyer.
\newblock {T}rivia{QA}: A large scale distantly supervised challenge dataset
  for reading comprehension.
\newblock In \emph{Proc. of ACL}, 2017{\natexlab{b}}.

\bibitem[Kaplan et~al.(2020)Kaplan, McCandlish, Henighan, Brown, Chess, Child,
  Gray, Radford, Wu, and Amodei]{Kaplan:etal:2020}
Jared Kaplan, Sam McCandlish, Tom Henighan, Tom~B. Brown, Benjamin Chess, Rewon
  Child, Scott Gray, Alec Radford, Jeffrey Wu, and Dario Amodei.
\newblock Scaling laws for neural language models, 2020.

\bibitem[Karpukhin et~al.(2020)Karpukhin, Oguz, Min, Lewis, Wu, Edunov, Chen,
  and Yih]{dpr}
Vladimir Karpukhin, Barlas Oguz, Sewon Min, Patrick Lewis, Ledell Wu, Sergey
  Edunov, Danqi Chen, and Wen-tau Yih.
\newblock Dense passage retrieval for open-domain question answering.
\newblock In \emph{Proc. of EMNLP}, 2020.

\bibitem[Khandelwal et~al.(2020)Khandelwal, Levy, Jurafsky, Zettlemoyer, and
  Lewis]{knn-lm}
Urvashi Khandelwal, Omer Levy, Dan Jurafsky, Luke Zettlemoyer, and Mike Lewis.
\newblock Generalization through memorization: Nearest neighbor language
  models.
\newblock In \emph{Proc. of ICLR}, 2020.

\bibitem[Kifer et~al.(2004)Kifer, Ben-David, and Gehrke]{kifer2004detecting}
Daniel Kifer, Shai Ben-David, and Johannes Gehrke.
\newblock Detecting change in data streams.
\newblock In \emph{VLDB}, volume~4, pages 180--191. Toronto, Canada, 2004.

\bibitem[Kirkpatrick et~al.(2017)Kirkpatrick, Pascanu, Rabinowitz, Veness,
  Desjardins, Rusu, Milan, Quan, Ramalho, Grabska-Barwinska, Hassabis, Clopath,
  Kumaran, and Hadsell]{Kirkpatrick3521}
James Kirkpatrick, Razvan Pascanu, Neil Rabinowitz, Joel Veness, Guillaume
  Desjardins, Andrei~A. Rusu, Kieran Milan, John Quan, Tiago Ramalho, Agnieszka
  Grabska-Barwinska, Demis Hassabis, Claudia Clopath, Dharshan Kumaran, and
  Raia Hadsell.
\newblock Overcoming catastrophic forgetting in neural networks.
\newblock \emph{Proceedings of the National Academy of Sciences}, 114\penalty0
  (13), 2017.

\bibitem[Kitaev et~al.(2020)Kitaev, Kaiser, and Levskaya]{kitaev_2020}
Nikita Kitaev, Lukasz Kaiser, and Anselm Levskaya.
\newblock Reformer: The efficient transformer.
\newblock In \emph{Proc. of ICLR}, 2020.

\bibitem[Krause et~al.(2018)Krause, Kahembwe, Murray, and Renals]{krause_2018}
Ben Krause, Emmanuel Kahembwe, Iain Murray, and Steve Renals.
\newblock Dynamic evaluation of neural sequence models.
\newblock In \emph{Proc. of ICML}, 2018.

\bibitem[Krause et~al.(2019)Krause, Kahembwe, Murray, and
  Renals]{Krause:etal:2019}
Ben Krause, Emmanuel Kahembwe, Iain Murray, and Steve Renals.
\newblock Dynamic evaluation of transformer language models.
\newblock \emph{arXiv preprint arXiv:1904.08378}, 2019.

\bibitem[Kruszewski et~al.(2020)Kruszewski, Sorodoc, and
  Mikolov]{Kruszewski:etal:2020}
Germán Kruszewski, Ionut-Teodor Sorodoc, and Tomas Mikolov.
\newblock Evaluating online continual learning with calm, 2020.

\bibitem[Kudo and Richardson(2018)]{Kudo:Richardson:2018}
Taku Kudo and John Richardson.
\newblock Sentencepiece: A simple and language independent subword tokenizer
  and detokenizer for neural text processing.
\newblock In \emph{Proceedings of the 2018 Conference on Empirical Methods in
  Natural Language Processing: System Demonstrations}, pages 66--71, 2018.

\bibitem[Kwiatkowski et~al.(2019)Kwiatkowski, Palomaki, Redfield, Collins,
  Parikh, Alberti, Epstein, Polosukhin, Devlin, Lee, Toutanova, Jones, Kelcey,
  Chang, Dai, Uszkoreit, Le, and Petrov]{natural-questions}
Tom Kwiatkowski, Jennimaria Palomaki, Olivia Redfield, Michael Collins, Ankur
  Parikh, Chris Alberti, Danielle Epstein, Illia Polosukhin, Jacob Devlin,
  Kenton Lee, Kristina Toutanova, Llion Jones, Matthew Kelcey, Ming-Wei Chang,
  Andrew~M. Dai, Jakob Uszkoreit, Quoc Le, and Slav Petrov.
\newblock Natural questions: A benchmark for question answering research.
\newblock \emph{TACL}, 7, March 2019.

\bibitem[Levenberg et~al.(2010)Levenberg, Callison-Burch, and
  Osborne]{levenberg_2010}
Abby Levenberg, Chris Callison-Burch, and Miles Osborne.
\newblock Stream-based translation models for statistical machine translation.
\newblock In \emph{Proc. of NAACL-HLT}, 2010.

\bibitem[Lewis et~al.(2020{\natexlab{a}})Lewis, Perez, Piktus, Petroni,
  Karpukhin, Goyal, K{\"u}ttler, Lewis, Yih, Rockt{\"a}schel,
  et~al.]{Lewis:etal:2020b}
Patrick Lewis, Ethan Perez, Aleksandara Piktus, Fabio Petroni, Vladimir
  Karpukhin, Naman Goyal, Heinrich K{\"u}ttler, Mike Lewis, Wen-tau Yih, Tim
  Rockt{\"a}schel, et~al.
\newblock Retrieval-augmented generation for knowledge-intensive nlp tasks.
\newblock \emph{arXiv preprint arXiv:2005.11401}, 2020{\natexlab{a}}.

\bibitem[Lewis et~al.(2020{\natexlab{b}})Lewis, Perez, Piktus, Petroni,
  Karpukhin, Goyal, K\"{u}ttler, Lewis, Yih, Rockt\"{a}schel, Riedel, and
  Kiela]{rag}
Patrick Lewis, Ethan Perez, Aleksandra Piktus, Fabio Petroni, Vladimir
  Karpukhin, Naman Goyal, Heinrich K\"{u}ttler, Mike Lewis, Wen-tau Yih, Tim
  Rockt\"{a}schel, Sebastian Riedel, and Douwe Kiela.
\newblock Retrieval-augmented generation for knowledge-intensive nlp tasks.
\newblock In H.~Larochelle, M.~Ranzato, R.~Hadsell, M.~F. Balcan, and H.~Lin,
  editors, \emph{Proc. of NeurIPS}, 2020{\natexlab{b}}.

\bibitem[Lewis et~al.(2020{\natexlab{c}})Lewis, Stenetorp, and
  Riedel]{Lewis:etal:2020}
Patrick Lewis, Pontus Stenetorp, and Sebastian Riedel.
\newblock Question and answer test-train overlap in open-domain question
  answering datasets.
\newblock \emph{arXiv preprint arXiv:2008.02637}, 2020{\natexlab{c}}.

\bibitem[Liu et~al.(2019)Liu, Ott, Goyal, Du, Joshi, Chen, Levy, Lewis,
  Zettlemoyer, and Stoyanov]{roberta}
Yinhan Liu, Myle Ott, Naman Goyal, Jingfei Du, Mandar Joshi, Danqi Chen, Omer
  Levy, Mike Lewis, Luke Zettlemoyer, and Veselin Stoyanov.
\newblock Roberta: {A} robustly optimized {BERT} pretraining approach.
\newblock \emph{CoRR}, abs/1907.11692, 2019.

\bibitem[Lopez-Paz and Ranzato(2017)]{GradientEpisodicMemory}
David Lopez-Paz and Marc'Aurelio Ranzato.
\newblock Gradient episodic memory for continual learning.
\newblock In \emph{Proc. of NeurIPS}, 2017.

\bibitem[Lu et~al.(2019)Lu, Liu, Dong, Gu, Gama, and
  Zhang]{concept_drift_review_2019}
Jie Lu, Anjin Liu, Fan Dong, Feng Gu, João Gama, and Guangquan Zhang.
\newblock Learning under concept drift: A review.
\newblock \emph{IEEE Transactions on Knowledge and Data Engineering},
  31\penalty0 (12), 2019.

\bibitem[Lukes and S{\o}gaard(2018)]{lukes-sogaard-2018-sentiment}
Jan Lukes and Anders S{\o}gaard.
\newblock Sentiment analysis under temporal shift.
\newblock In \emph{Proc. of WASSA}, 2018.

\bibitem[Mccloskey and Cohen(1989)]{mccloskey:catastrophic}
Michael Mccloskey and Neil~J. Cohen.
\newblock Catastrophic interference in connectionist networks: {T}he sequential
  learning problem.
\newblock \emph{The Psychology of Learning and Motivation}, 24, 1989.

\bibitem[Mielke et~al.(2019)Mielke, Cotterell, Gorman, Roark, and
  Eisner]{Mielke:etal:2019}
Sabrina~J. Mielke, Ryan Cotterell, Kyle Gorman, Brian Roark, and Jason Eisner.
\newblock What kind of language is hard to language-model?
\newblock In \emph{Proc. of ACL}, 2019.

\bibitem[Mikolov et~al.(2010)Mikolov, Karafiát, Burget, Cernocký, and
  Khudanpur]{mikolov_etal_2010}
Tomas Mikolov, Martin Karafiát, Lukás Burget, Jan Cernocký, and Sanjeev
  Khudanpur.
\newblock Recurrent neural network based language model.
\newblock In \emph{Proc. of INTERSPEECH}, 2010.

\bibitem[Mitchell et~al.(2015)Mitchell, Cohen, Hruschka, Talukdar, Betteridge,
  Carlson, Dalvi, Gardner, Kisiel, Krishnamurthy, Lao, Mazaitis, Mohamed,
  Nakashole, Platanios, Ritter, Samadi, Settles, Wang, Wijaya, Gupta, Chen,
  Saparov, Greaves, and Welling]{mitchell_2015}
T.~Mitchell, W.~Cohen, E.~Hruschka, P.~Talukdar, J.~Betteridge, A.~Carlson,
  B.~Dalvi, M.~Gardner, B.~Kisiel, J.~Krishnamurthy, N.~Lao, K.~Mazaitis,
  T.~Mohamed, N.~Nakashole, E.~Platanios, A.~Ritter, M.~Samadi, B.~Settles,
  R.~Wang, D.~Wijaya, A.~Gupta, X.~Chen, A.~Saparov, M.~Greaves, and
  J.~Welling.
\newblock Never-ending learning.
\newblock In \emph{Proc. of AAAI}, 2015.

\bibitem[Nie et~al.(2020)Nie, Williams, Dinan, Bansal, Weston, and
  Kiela]{Nie:etal:2019}
Yixin Nie, Adina Williams, Emily Dinan, Mohit Bansal, Jason Weston, and Douwe
  Kiela.
\newblock Adversarial {NLI}: A new benchmark for natural language
  understanding.
\newblock In \emph{Proc. of ACL}, 2020.

\bibitem[Oren et~al.(2019)Oren, Sagawa, Hashimoto, and Liang]{Oren:etal:2019}
Yonatan Oren, Shiori Sagawa, Tatsunori Hashimoto, and Percy Liang.
\newblock Distributionally robust language modeling.
\newblock In \emph{Proc. of EMNLP-IJCNLP}, 2019.

\bibitem[Osborne et~al.(2014)Osborne, Lall, and
  Van~Durme]{Osborne:VanDurme:2014}
Miles Osborne, Ashwin Lall, and Benjamin Van~Durme.
\newblock Exponential reservoir sampling for streaming language models.
\newblock In \emph{Proc. of ACL}, 2014.

\bibitem[Paperno et~al.(2016)Paperno, Kruszewski, Lazaridou, Pham, Bernardi,
  Pezzelle, Baroni, Boleda, and Fern{\'a}ndez]{lambada}
Denis Paperno, Germ{\'a}n Kruszewski, Angeliki Lazaridou, Ngoc~Quan Pham,
  Raffaella Bernardi, Sandro Pezzelle, Marco Baroni, Gemma Boleda, and Raquel
  Fern{\'a}ndez.
\newblock The {LAMBADA} dataset: Word prediction requiring a broad discourse
  context.
\newblock In \emph{Proc. of ACL}, 2016.

\bibitem[Patterson et~al.(2021)Patterson, Gonzalez, Le, Liang, Munguia,
  Rothchild, So, Texier, and Dean]{carbon-impact-2021}
David~A. Patterson, Joseph Gonzalez, Quoc~V. Le, Chen Liang, Lluis{-}Miquel
  Munguia, Daniel Rothchild, David~R. So, Maud Texier, and Jeff Dean.
\newblock Carbon emissions and large neural network training.
\newblock \emph{CoRR}, abs/2104.10350, 2021.

\bibitem[Potts et~al.(2020)Potts, Wu, Geiger, and Kiela]{Potts:etal:2020}
Christopher Potts, Zhengxuan Wu, Atticus Geiger, and Douwe Kiela.
\newblock Dynasent: A dynamic benchmark for sentiment analysis.
\newblock \emph{arXiv preprint arXiv:2012.15349}, 2020.

\bibitem[Radford et~al.(2019)Radford, Wu, Child, Luan, Amodei, and
  Sutskever]{Radford:etal:2019}
Alec Radford, Jeffrey Wu, Rewon Child, David Luan, Dario Amodei, and Ilya
  Sutskever.
\newblock Language models are unsupervised multitask learners.
\newblock In \emph{Technical report, OpenAI.}, 2019.

\bibitem[Rijhwani and
  Preotiuc-Pietro(2020)]{rijhwani-preotiuc-pietro-2020-temporally}
Shruti Rijhwani and Daniel Preotiuc-Pietro.
\newblock Temporally-informed analysis of named entity recognition.
\newblock In \emph{Proc. of ACL}, 2020.

\bibitem[Rosenfeld and Erk(2018)]{rosenfeld2018deep}
Alex Rosenfeld and Katrin Erk.
\newblock Deep neural models of semantic shift.
\newblock In \emph{Proc. of NAACL-HLT}, 2018.

\bibitem[Rusu et~al.(2016)Rusu, Rabinowitz, Desjardins, Soyer, Kirkpatrick,
  Kavukcuoglu, Pascanu, and Hadsell]{rusu_etal_2016}
Andrei~A. Rusu, Neil~C. Rabinowitz, Guillaume Desjardins, Hubert Soyer, James
  Kirkpatrick, Koray Kavukcuoglu, Razvan Pascanu, and Raia Hadsell.
\newblock Progressive neural networks.
\newblock \emph{CoRR}, abs/1606.04671, 2016.

\bibitem[Sinitsin et~al.(2020)Sinitsin, Plokhotnyuk, Pyrkin, Popov, and
  Babenko]{Sinitsin:etal:2020}
Anton Sinitsin, Vsevolod Plokhotnyuk, Dmitriy Pyrkin, Sergei Popov, and Artem
  Babenko.
\newblock Editable neural networks.
\newblock \emph{arXiv preprint arXiv:2004.00345}, 2020.

\bibitem[S{\o}gaard et~al.(2021)S{\o}gaard, Ebert, Bastings, and
  Filippova]{Sogaard:etal:2020}
Anders S{\o}gaard, Sebastian Ebert, Jasmijn Bastings, and Katja Filippova.
\newblock We need to talk about random splits.
\newblock In \emph{Proc. of EACL}, 2021.

\bibitem[Strubell et~al.(2019)Strubell, Ganesh, and McCallum]{strubell-2019}
Emma Strubell, Ananya Ganesh, and Andrew McCallum.
\newblock Energy and policy considerations for deep learning in {NLP}.
\newblock In \emph{Proc. of ACL}, 2019.

\bibitem[Sun et~al.(2020)Sun, Ho, and Lee]{Sun2020lamal}
Fan-Keng Sun, Cheng-Hao Ho, and Hung-Yi Lee.
\newblock {LAMOL: LAnguage MOdeling for Lifelong Language Learning}.
\newblock In \emph{Proc.of ICLR}, 2020.

\bibitem[Szymanski(2017)]{Szymanski2017analogies}
Terrence Szymanski.
\newblock {Temporal Word Analogies: Identifying Lexical Replacement with
  Diachronic Word Embeddings}.
\newblock In \emph{Proc. of ACL}, 2017.

\bibitem[Thorne and Vlachos(2018)]{thorne-vlachos-2018-automated}
James Thorne and Andreas Vlachos.
\newblock Automated fact checking: Task formulations, methods and future
  directions.
\newblock In \emph{Proc. of ICCL}, 2018.

\bibitem[Thrun and Mitchell(1995)]{THRUN199525}
Sebastian Thrun and Tom~M. Mitchell.
\newblock Lifelong robot learning.
\newblock \emph{Robotics and Autonomous Systems}, 15\penalty0 (1), 1995.

\bibitem[Trischler et~al.(2017)Trischler, Wang, Yuan, Harris, Sordoni, Bachman,
  and Suleman]{trischler-etal-2017-newsqa}
Adam Trischler, Tong Wang, Xingdi Yuan, Justin Harris, Alessandro Sordoni,
  Philip Bachman, and Kaheer Suleman.
\newblock {N}ews{QA}: A machine comprehension dataset.
\newblock In \emph{Proceedings of the 2nd Workshop on Representation Learning
  for {NLP}}, pages 191--200, Vancouver, Canada, August 2017. Association for
  Computational Linguistics.
\newblock \doi{10.18653/v1/W17-2623}.
\newblock URL \url{https://www.aclweb.org/anthology/W17-2623}.

\bibitem[Vaswani et~al.(2017)Vaswani, Shazeer, Parmar, Uszkoreit, Jones, Gomez,
  Kaiser, and Polosukhin]{Vaswani:etal:2017}
Ashish Vaswani, Noam Shazeer, Niki Parmar, Jakob Uszkoreit, Llion Jones,
  Aidan~N Gomez, \L~ukasz Kaiser, and Illia Polosukhin.
\newblock Attention is all you need.
\newblock In \emph{Proc. of NeurIPS}, 2017.

\bibitem[Vinyals et~al.(2015)Vinyals, Fortunato, and Jaitly]{vinyals_2015}
Oriol Vinyals, Meire Fortunato, and Navdeep Jaitly.
\newblock Pointer networks.
\newblock In \emph{Proc. of NeurIPS}, 2015.

\bibitem[Wang et~al.(2008)Wang, Blei, and Heckerman]{Wang:etal:2008}
Chong Wang, David Blei, and David Heckerman.
\newblock Continuous time dynamic topic models.
\newblock In \emph{Proc. of UAI}, 2008.

\bibitem[Widmer and Kubat(1996)]{concept_drift_widmer_1996}
Gerhard Widmer and Miroslav Kubat.
\newblock Learning in the presence of concept drift and hidden contexts.
\newblock \emph{Machine Learning}, 23\penalty0 (1), 1996.

\bibitem[Yin et~al.(2018)Yin, Sachidananda, and Prabhakar]{yin2018global}
Zi~Yin, Vin Sachidananda, and Balaji Prabhakar.
\newblock The global anchor method for quantifying linguistic shifts and domain
  adaptation.
\newblock \emph{Proc. of NeurIPS}, 2018.

\bibitem[Yogatama et~al.(2014)Yogatama, Wang, Routledge, Smith, and
  Xing]{Yogatama:etal:2014}
Dani Yogatama, Chong Wang, Bryan~R. Routledge, Noah~A. Smith, and Eric~P. Xing.
\newblock Dynamic language models for streaming text.
\newblock \emph{Transactions of Association of Computational Linguistics},
  2014.

\bibitem[Yogatama et~al.(2021)Yogatama, de~Masson~d'Autume, and
  Kong]{Yogatama:etal:2021}
Dani Yogatama, Cyprien de~Masson~d'Autume, and Lingpeng Kong.
\newblock Adaptive semiparametric language models.
\newblock \emph{Transactions of Association of Computational Linguistics},
  2021.

\bibitem[Zellers et~al.(2019)Zellers, Holtzman, Rashkin, Bisk, Farhadi,
  Roesner, and Choi]{zellers_etal_2019}
Rowan Zellers, Ari Holtzman, Hannah Rashkin, Yonatan Bisk, Ali Farhadi,
  Franziska Roesner, and Yejin Choi.
\newblock Defending against neural fake news.
\newblock In \emph{Proc. of NeurIPS}, 2019.

\bibitem[Zhu et~al.(2020)Zhu, Rawat, Zaheer, Bhojanapalli, Li, Yu, and
  Kumar]{Zhu:etal:2020}
Chen Zhu, Ankit~Singh Rawat, Manzil Zaheer, Srinadh Bhojanapalli, Daliang Li,
  Felix Yu, and Sanjiv Kumar.
\newblock Modifying memories in transformer models.
\newblock \emph{arXiv preprint arXiv:2012.00363}, 2020.

\end{thebibliography}

\section*{Checklist}

\begin{enumerate}

\item For all authors...
\begin{enumerate}
  \item Do the main claims made in the abstract and introduction accurately reflect the paper's contributions and scope?
    \answerYes{}
  \item Did you describe the limitations of your work?
    \answerYes{See the ``Limitation'' part of Section \ref{sec:discussion}.}
  \item Did you discuss any potential negative societal impacts of your work?
    \answerYes{We work with large-scale language models. In the paper, we have outlined the broader societal impact of our work, although other risks that stem from language modelling research on large amounts of data may also apply to our work \citep{Bender:etal:2021}.}
  \item Have you read the ethics review guidelines and ensured that your paper conforms to them?
    \answerYes{We have outlined the potential negative social impacts of our work in \S\ref{sec:societal_impact}, and additionally in the answer to Checklist 1.(c) above.}
\end{enumerate}

\item If you are including theoretical results...
\begin{enumerate}
  \item Did you state the full set of assumptions of all theoretical results?
    \answerNA{This paper does not include theoretical results.}
	\item Did you include complete proofs of all theoretical results?
    \answerNA{}
\end{enumerate}

\item If you ran experiments...
\begin{enumerate}
  \item Did you include the code, data, and instructions needed to reproduce the main experimental results (either in the supplemental material or as a URL)?
    \answerYes{All dataset details and preprocessing steps are described in Section \ref{sec:data}. While we do not release the code, the experiment can be repeated with publicly available Transformer implementations. The dataset splits are released publicly.}
  \item Did you specify all the training details (e.g., data splits, hyperparameters, how they were chosen)?
     \answerYes{The dataset and splits are described in Section \ref{sec:data}, and are released publicly. The hyper-parameters and model details are described in Section \ref{sec:model}.}
	\item Did you report error bars (e.g., with respect to the random seed after running experiments multiple times)?
    \answerYes{We have conducted significance testing and also tested the robustness of our findings by replicating our experiments in different configurations (e.g., with larger model sizes, in different languages, in different datasets, and in different yearly splits of our datasets).}
	\item Did you include the total amount of compute and the type of resources used (e.g., type of GPUs, internal cluster, or cloud provider)?
    \answerYes{To train and evaluate the models, including hyperparameter optimization, we used approximately 186,000 TPU hours. In each experiment, we used 32 TPUs for training and 1 TPU for evaluation.}
\end{enumerate}

\item If you are using existing assets (e.g., code, data, models) or curating/releasing new assets...
\begin{enumerate}
  \item If your work uses existing assets, did you cite the creators?
    \answerYes{We build our data using existing datasets, which we cite in footnote 4.}
  \item Did you mention the license of the assets?
    \answerNA{We release dataset splits as lists of identifiers pointing to original datasets. The licenses of the original datasets apply.}
  \item Did you include any new assets either in the supplemental material or as a URL?
    \answerYes{We are providing the URL to the publicly released resources.}
  \item Did you discuss whether and how consent was obtained from people whose data you're using/curating?
    \answerNA{We build all our datasets through publicly available datasets.}
  \item Did you discuss whether the data you are using/curating contains personally identifiable information or offensive content?
    \answerNo{We work with publicly available datasets that have been used in prior work. An analysis of whether, and to what extent, personally identifiable information can be extracted from these publicly available datasets are beyond the scope of this work.}
\end{enumerate}

\item If you used crowdsourcing or conducted research with human subjects...
\begin{enumerate}
  \item Did you include the full text of instructions given to participants and screenshots, if applicable?
    \answerNA{This work does not include any crowd-sourcing or research with human subjects.}
  \item Did you describe any potential participant risks, with links to Institutional Review Board (IRB) approvals, if applicable?
    \answerNA{}
  \item Did you include the estimated hourly wage paid to participants and the total amount spent on participant compensation?
    \answerNA{}
\end{enumerate}

\end{enumerate}


\appendix
\clearpage

\section{How General Are These Findings?}

\subsection{The effect of outdated models persists beyond the 2018/2019 test period.}
\label{sec:rolling}

We test whether the temporal degradation trends we observe in \S\ref{sec:experiment1} are \emph{not} an artifact of some particularity of the chosen test period (i.e., $Yr1=2018$ and $Yr2=2019$). We design new test sets by shifting $Yr1$ and $Yr2$ in increments of one year towards the past, for a total of five such test sets. Following \S\ref{sec:setup}, we derive different  \timestratified$^{Yr1, Yr2}$ and \control$^{Yr1, Yr2}$~training and validation splits. 

\begin{figure}[h]
 \centering
    \centering
    \includegraphics[width=9cm,height=5cm,keepaspectratio]{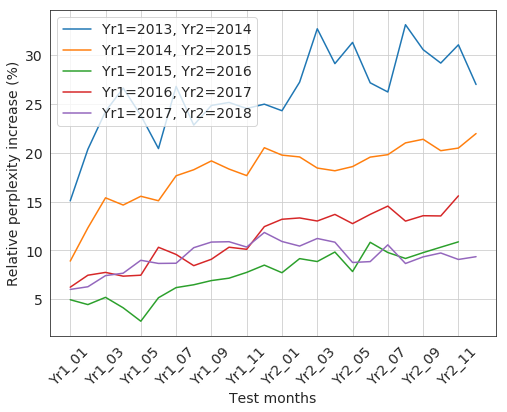}
\caption{Relative perplexity increase of \timestratified$^{Yr1, Yr2}$~over \control$^{Yr1, Yr2}$~models.}
 \label{fig:rolling}
\end{figure}

Note that each \timestratified$^{Yr1, Yr2}$ and \control$^{Yr1, Yr2}$ setups are: (i) Trained on the same training data sizes, and (ii) evaluated on the same test set covering $Yr1$ and $Yr2$. Fig.~\ref{fig:rolling} shows similar temporal degradation across all testing years.

\subsection{The effect of outdated models persists beyond the two-year gap.}
\label{sec:increasing}

\begin{figure}[h]
 \centering
    \centering
    \includegraphics[width=9cm,height=5cm,keepaspectratio]{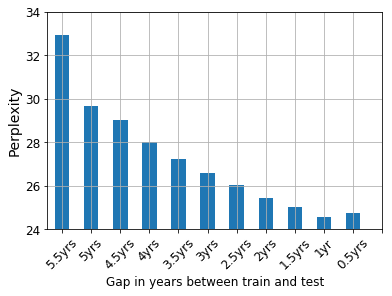}
\caption{Perplexity of models trained with data from different time periods, with increasingly larger gaps from the 2018-2019 test set period.}
\label{fig:increasing}
\end{figure}

For this experiment, we keep the same 2018-2019 test set introduced in \S\ref{sec:setup}, and train models with training data from different time periods with increasingly larger gaps from the 2018-2019 evaluation period, controlling so that all training data sizes are identical across different years. More concretely, the most up-to-date model covers the same time period as the original \timestratified~model, and we ``push'' the training period back with 6-month increments, up to September 2012, for a total of 11 training sets---each of the same size---used to train 11 models. Fig.~\ref{fig:increasing} shows that the perplexity deterioration continues to grow in response to larger gaps between the training and test periods.

\subsection{The effect of outdated models persists beyond English: A German study.}
\label{sec:german}

We test whether the temporal degradation trend is a generalizable pattern that holds across languages. We use the German subset of \wmt, apply the same pre-processing steps as \S\ref{sec:data}, follow the same experimental setup as \S\ref{sec:setup}, and train two Transformer-XL models on \timestratified$^{de}$ and \control$^{de}$ setups, achieving 30.87 and 26.79 respective test perplexities. These perplexities are indeed higher than the ones in Table~\ref{tab:overall_perplexity}---a consistent pattern with prior findings on the difficulty of modelling German \citep{Mielke:etal:2019}. Nevertheless, we still see the exact same pattern where the stale \timestratified$^{de}$ model performs worse than the \control$^{de}$ one (a substantial 15.23\% relative increase). Moreover, similar to the English experiment, the model degrades more as the gap between the training and test period increases---an  effect particularly pronounced for proper nouns and for words that are broken down by the \timestratified$^{de}$ tokenizer into more tokens. 

\begin{figure}[h]
 \centering
    \centering
    \includegraphics[width=9cm,height=5cm,keepaspectratio]{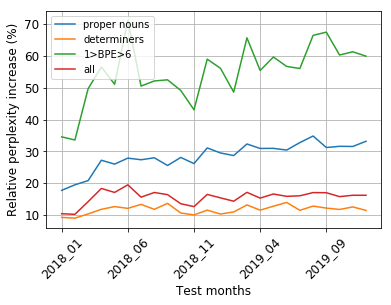}
    \label{fig:german}
\caption{For the experiments on German, the relative increase of perplexity of the \timestratified$^{de}$ model over its \control$^{de}$ counterpart.}
\end{figure}

\section{Dynamic evaluation}
\label{sec:appendix_dynamic}
Here we more formally describe dynamic evaluation, which we apply to the \timestratified~model, and outline some of the hyper-parameter choices used for our dynamic evaluation experiments (\S\ref{sec:solution}). Let $\{D^{(1)}, D^{(2)}, \cdots, D^{(N)}\}$ be a collection of $N$ chronologically-ordered test documents, where $D^{(t-1)}$ was published before $D^{(t)}$, and $D^{(1)}$ was our first test document in the 2018-2019 evaluation period (\S\ref{sec:data}). Each test document $D^{(t)}$ consists of $M=|D^{(t)}|$ tokens $\mathbf{x}^{(t)}=x^{(t)}_1, x^{(t)}_2, \cdots, x^{(t)}_{M}$. Furthermore, let $\boldsymbol{\theta}_1$ be the set of Transformer-XL model parameters (\S\ref{sec:model}) \emph{after} training on documents from the pre-2018 training period (\timestratified~setup; \S\ref{sec:data}), and \emph{before} any dynamic evaluation is applied. 

The loss of the Transformer-XL model with respect to a test document $D^{(t)}$ is computed as follows:
\begin{align}
    \ell(D^{(t)}; \boldsymbol{\theta}_{t}) = \log  p_{\boldsymbol{\theta}_{t}}(\mathbf{x}^{(t)}) = \log \left( \prod_{i=1}^M p_{\boldsymbol{\theta}_{t}}(x^{(t)}_i \, | \, \mathbf{x}^{(t)}_{<i}) \right) = \sum_{i=1}^M \log p_{\boldsymbol{\theta}_{t}}(x^{(t)}_i \, | \, \mathbf{x}^{(t)}_{<i}),
\end{align}
where $\mathbf{x}^{(t)}_{<i}$ denotes tokens $x^{(t)}_1,x^{(t)}_2,\cdots,x^{(t)}_{i-1}$ in the test document $D^{(t)}$ that precede $x^{(t)}_i$.

In dynamic evaluation \citep{mikolov_etal_2010,graves_2013,krause_2018,Krause:etal:2019}, we dynamically update the Transformer-XL model parameters using gradient descent, based on the knowledge contained in the test documents that had been seen so far. More formally,
\begin{align}
    \boldsymbol{\theta}_{t+1} \leftarrow \boldsymbol{\theta}_{t} - \alpha \, \nabla_{\boldsymbol{\theta}_{t}} \, \ell(D^{(t)}; \boldsymbol{\theta}_{t}), \label{eq:dynamic_eval}
\end{align}
where $\alpha$ denotes the dynamic evaluation learning rate, and $\nabla_{\boldsymbol{\theta}_{t}} \, \ell(D^{(t)}; \boldsymbol{\theta}_{t})$ denotes the gradient of the model parameters with respect to the model's loss for the current document $\ell(D^{(t)}; \boldsymbol{\theta}_{t})$.

This procedure means that the model parameters $\boldsymbol{\theta}_{t}$, which we use to evaluate the model on the current test document $D^{(t)}$, \emph{already encodes} knowledge from previous test documents $D^{(1)}, D^{(2)}, \cdots, D^{(t-1)}$, in addition to the knowledge learnt from the training set. This in turn enables the model to learn about new information that emerges or becomes more salient during the evaluation period (e.g. ``COVID-19'' in late-2019), which is then stored in the model parameters, and reuse such information for better prediction of subsequent test documents. In practice, our implementation of dynamic evaluation differs from Eq.~\ref{eq:dynamic_eval} in two ways: (i) We perform $K$ steps of gradient descent for each document, rather than only one step, where $K$ is tuned on the validation set; and (ii) we perform the gradient updates for a batch of contiguous tokens (e.g. 512), which means that documents that are longer than the batch size will have more than one parameter update.

\paragraph{Contrast with non-dynamic evaluation.} When dynamic evaluation is not applied, $\boldsymbol{\theta}_{t} = \boldsymbol{\theta}_{t-1} = \boldsymbol{\theta}_{1}$. This means that the same model parameters $\boldsymbol{\theta}_{1}$ (i.e. model parameters after training on the training documents---\emph{without} updating the models' knowledge on the observed test documents) are used to predict all test documents, risking the model becoming outdated in-between retraining cycles.

\paragraph{Dynamic evaluation hyper-parameters.} We use the following learning rates (\wmt: 5e-5, \customnews:5e-4, \arxiv: 1e-3), which are tuned on the validation set spanning three months, whereas the test set spans two years. We leave the question of choosing a learning rate with an optimal trade-off between adaptation speed and stability of updates without \emph{a priori} knowledge of the evaluation period to future work.

\subsection{Dynamic Evaluation and Catastrophic Forgetting}

We design an experiment to assess whether updating a model on present data using dynamic evaluation leads to catastrophic forgetting of the past data. To assess this, we report the performance of the two models, i.e., the one trained until 2017 and the one updated up to 2019, on a test set derived from the initial training data of the model covering the years up to the year from which we started performing dynamic evaluation (i.e., 2007-2017). In addition, we also report the results on the 2018-2019 test set which were presented in Section~\ref{sec:solution}.

\begin{figure}[h]
 \centering
    \centering
    \includegraphics[width=9cm,height=5cm,keepaspectratio]{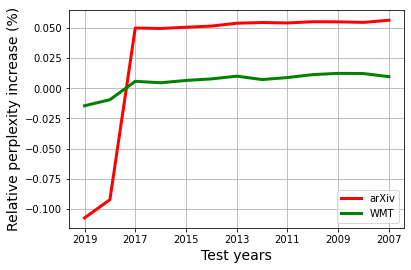}
\caption{Catastrophic forgetting as measures in terms of relative perplexity increase when comparing the models updated with dynamic evaluation against the models that have been trained with data up to 2017. The x-axis presents the years in a reverse chronological order.}
\label{fig:catastrophic_forgetting}
\end{figure}

Figure~\ref{fig:catastrophic_forgetting} presents the results for \wmt~and\arxiv. For both datasets we observe that as we move towards the past, the perplexity of the model updated with dynamic evaluation increases. As such, while the updated model outperforms the outdated model for the recent 2018 and 2019 years, the same model performs increasingly worse on the past years, as indicated by the gentle upward slope from 2017 and onwards.

\section{Example Question-Answer Pairs}
\label{sec:appendix_downstream}

\subsection{Examples of closed-book QA on synthetic questions on government officials}

\textbf{Question:} Who was the governor in Texas on 5 September 2019? \textbf{Answer:} Greg Abbott

\textbf{Question:} Who was the prime minister in Canada on 8 June 2019? \textbf{Answer:} Justin Trudeau

\textbf{Question:} Who was the president in Portugal on 30 May 2019?	\textbf{Answer:} Marcelo Rebelo de Sousa

\subsection{Examples of reading comprehension on NewsQA}

\textbf{Document:}
England international footballer Steven Gerrard was found not guilty of affray by a court in his home city on Friday. England international Steven Gerrard was cleared by a court in Liverpool of affray. The jury at Liverpool Crown Court took a little over an hour to clear Gerrard of charges relating to a fracas in a nightclub bar in the north-western of England city on December 29 of last year. They accepted the Liverpool captain\'s version that he acted in self defense in punching businessman Marcus McGhee. The 29-year-old was the only one of the seven defendants in the case to be cleared after an incident which was described by judge Henry Globe as an "explosion of violence." Gerrard spoke of his relief outside the court. "Can I just say how pleased I am with today\'s verdict," he said. "I\'m glad to put this case behind me and I am really looking forward to the season ahead and concentrating on my football now. "I would just like to say a big thank you to my legal team and to my friends and family and everyone at Liverpool football club for supporting me." His comments were met with a round of applause from a large group of fans of the Premier League club who had gathered outside the court, before he was ushered away. Gerrard was celebrating in the Lounge Inn in Southport, a suburb of Liverpool, after scoring twice his team\'s 5-1 win at Newcastle which took them to the top of the Premier League. Video footage, which was available to the court, showed.

\textbf{Question:} Who was cleared by a Liverpool court? \textbf{Answer:} Steven Gerrard

\textbf{Document:} CNN affiliates report on where job seekers are finding work across the country and how those looking for employment are coping with the situation. A census employee poses with the new handheld device field workers will use for the 2010 count. (CNN) -- The nation will take roll call in 2010 and the federal government is giving the states money to hire thousands of census workers. Officials in Colorado say they may hire as many as 8,000 workers for positions that last between 10 weeks and one year. Cathy Illian says the bureau has already hired 800 people in the Denver area. The organization will also post open positions in early April. Some jobs pay as much as \$28.75 an hour. Read the story on KMGH. In Idaho, Dave Mulvihill, manager of the state\'s census bureau, said the organization will hire 1,200 workers. He has plenty of job searchers to choose from. "We\'ve had applications from approximately 7,300 people across the state," he told CNN affiliate KIVI. Read the full report on census jobs. The office is holding off on taking any more applications until fall. The Alabama census bureau is preparing to hire between 1,000 and 1,500 workers. "We need workers so we can get good addresses [to] send the questionnaires out so we can get a good response," state census bureau official Darryl Lee told TV Alabama in Birmingham. Census officials point out that an accurate count of U.S. citizens helps the government figure out how much funding to give each state for federally sponsored programs. Read the ABC 33/40 story Northeast: Rhode Island strip club. 

\textbf{Question:} Census bureaus are hiring people from where? \textbf{Answer:} Denver area
\end{document}